\documentclass{article}

\usepackage{microtype}
\usepackage{graphicx}
\usepackage{subcaption}
\usepackage{booktabs} 

\usepackage{hyperref}

\usepackage{cuted}

\usepackage[preprint]{icml2026}
\makeatletter
\let\ICMLorigPrintAffiliationsAndNotice\printAffiliationsAndNotice
\renewcommand{\printAffiliationsAndNotice}[1]{%
  \begingroup
  \renewcommand{\footnotetext}[2][]{}
  \let\@footnotetext\@gobble
  \ICMLorigPrintAffiliationsAndNotice{}%
  \endgroup
}
\makeatother

\usepackage{amsmath}
\usepackage{amssymb}
\usepackage{mathtools}
\usepackage{amsthm}
\usepackage{marvosym}

\usepackage[capitalize,noabbrev]{cleveref}

\theoremstyle{plain}

\theoremstyle{definition}

\theoremstyle{remark}

\newcommand{\tablestyle}[2]{\setlength{\tabcolsep}
{#1}\renewcommand{\arraystretch}{#2}\centering\footnotesize}
\usepackage{multirow}
\usepackage[table]{xcolor} 
\usepackage{colortbl}      
\usepackage{pifont}        
\usepackage{amsmath}       
\usepackage{amssymb}       
\usepackage{graphicx}      

\definecolor{mygreen}{HTML}{2ECC71} 
\definecolor{myred}{HTML}{E74C3C}     
\definecolor{myblue}{HTML}{3498DB}    
\newcommand{\mj}{$\mathcal{J}$\xspace}
\newcommand{\mf}{$\mathcal{F}$\xspace}
\newcommand{\mjf}{$\mathcal{J}\&\mathcal{F}$\xspace}
\newcommand{\mg}{$\mathcal{G}$}

\newcommand{\cmark}{\textcolor{mygreen}{\ding{51}}}
\newcommand{\xmark}{\textcolor{myred}{\ding{55}}}

\usepackage{xspace}

\def\modelname{G\textsuperscript{2}TAM\xspace}

\definecolor{darkgreen}{rgb}{0.0,0.5,0.0}
\definecolor{cvprgreen}{RGB}{34, 139, 34}   
\definecolor{cvprgray}{RGB}{105, 105, 105}  
\definecolor{cvprorange}{RGB}{252, 211, 130} 
\definecolor{shadecolor}{RGB}{237,237,237}

\icmltitlerunning{Geometry Grounded Track Anything Model}

\begin{document}

\twocolumn[
  \icmltitle{\modelname: Geometry Grounded Track Anything Model}




  \begin{center}
    \textbf{Chenming Zhu}$^{1,2}$, \textbf{Peizhou Cao}$^{2,3}$, \textbf{Jingli Lin}$^{2,4}$, \textbf{Wenbo Hu}$^{2,5}$, \textbf{Yunlong Ran}$^{2,6}$, \\ \textbf{Jiangmiao Pang}$^2$, 
    \textbf{Tai Wang}$^2$, \textbf{Xihui Liu}$^{1}\textsuperscript{\Letter}$
  \end{center}
  
  \begin{center}
    \footnotesize
    $^1$ HKU \quad
    $^2$ Shanghai AI Lab \quad
    $^3$ BUAA \quad
    $^4$ SJTU \quad
    $^5$ UCLA \quad
    $^6$ ZJU \quad
    \\
    \vspace{0.02in}
    $\textsuperscript{\Letter}$ Corresponding Authors \quad
  \end{center}
  \vspace{0.08in}

  {\centering \tt\small \url{https://zcmax.github.io/projects/G2TAM/} \par}

  \vskip 0.3in
]

\printAffiliationsAndNotice{}  

\begin{strip}
    \centering
    \includegraphics[width=1\textwidth]{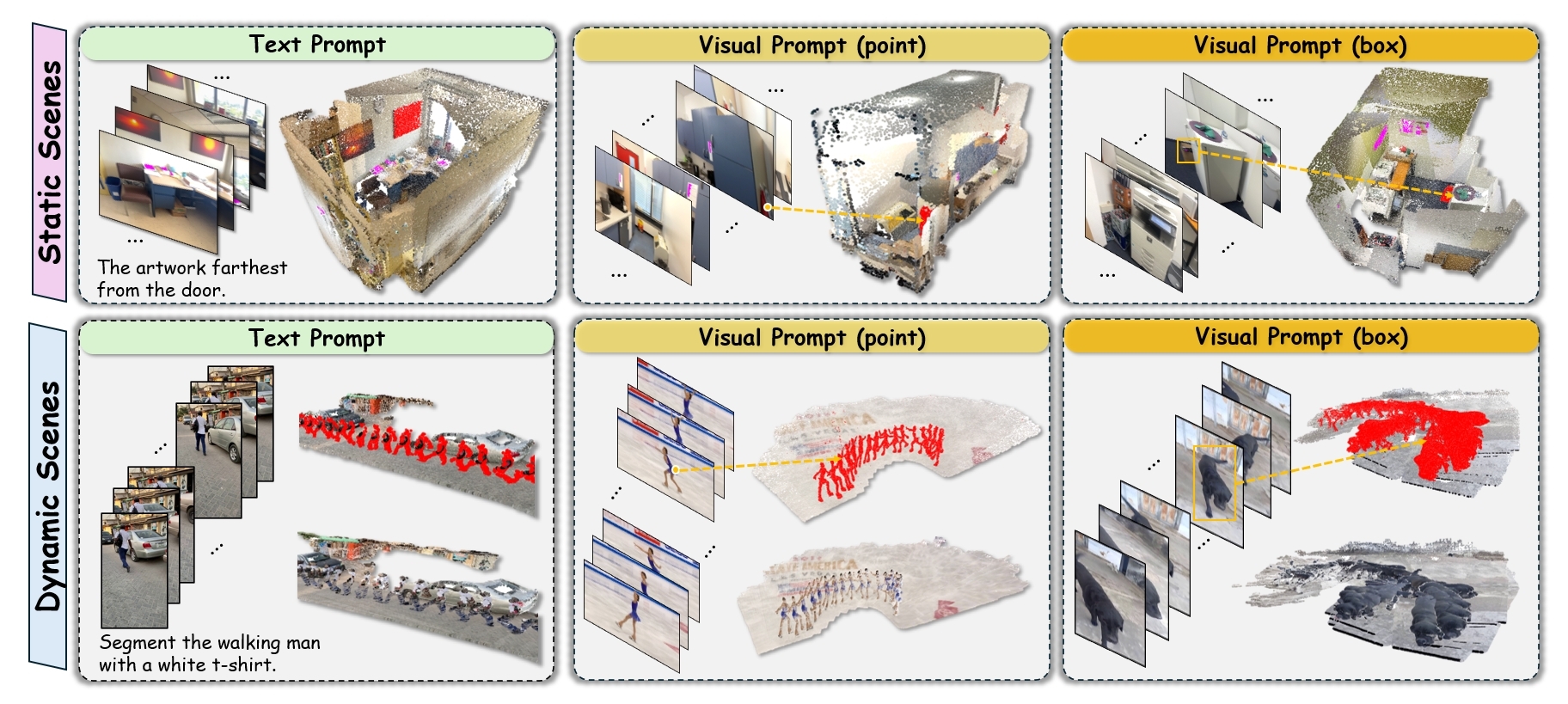}
    \captionof{figure}{
        Given unordered images or video inputs, \textbf{\modelname} supports various prompts—including text, point, and box prompts—to perform joint 3D reconstruction and spatial–temporal consistent instance segmentation, enabling promptable instance tracking in 3D space. 
    }
    \label{fig:new_teaser}
\end{strip}


\begin{abstract}
Human spatial understanding arises from jointly perceiving geometry and semantics, enabling consistent object identification and localization across viewpoints and time.  
Current video segmentation models depend on explicit object appearance memory banks for instance tracking, yet they remain vulnerable to large viewpoint changes and long-term occlusions.  
Leveraging the spatial consistency afforded by modern feed-forward 3D reconstruction models, we propose the \textbf{Geometry Grounded Tracking Anything Model (\modelname)}, a unified framework for promptable instance tracking in 3D using only unordered RGB images or videos. \modelname employs spatially aligned geometric representations as implicit memory, ensuring stable instance identity and localization across frames and views.  
At its core is a cross-modal spatial encoder that integrates visual and textual prompts into a shared geometric space, enabling end-to-end spatial reconstruction and instance-consistent mask prediction.  
To support training and evaluation, we construct \textbf{InsTrack}, a large-scale dataset with a dedicated validation split for benchmarking.  
Extensive experiments show that \modelname delivers strong cross-view consistency, promptable instance spatial tracking, video object segmentation and spatial reconstruction, establishing a foundation for interactive, geometry-grounded spatial reasoning.
\end{abstract}

\section{Introduction}

Humans recognize and track objects not by relying on appearance alone, but by integrating visual cues with a stable sense of 3D structure.
Even when an object’s appearance changes due to lighting, deformation, or drastic viewpoint shifts, humans maintain identity by grounding perception in a coherent spatial understanding of the scene.
This complementary use of appearance and geometry enables reliable tracking in both static environments and dynamic, motion-heavy scenarios.

Current video object segmentation (VOS) methods~\cite{ravi2024sam2, kirillov2023segment, li2023semantic} typically rely on appearance matching and explicit memory banks to perform temporal-consistent video object tracking, which often falter under drastic viewpoint shifts or long-term occlusions. While recent efforts~\cite{LMK, 3dtracking} attempt to incorporate 3D geometry information into egocentric tracking to achieve object permanence even when objects are out-of-view. These methods typically follow a ``Tracking-by-Mapping'' paradigm, which performs post-hoc matching by projecting explicit 2D tracking or detection results into 3D space via off-the-shelf pose and depth estimators. Such multi-stage pipelines may inherently suffer from error propagation between disconnected 2D and 3D modules and are strictly limited to scenarios where accurate geometry information is available.

With the recent progress in feed-forward 3D reconstruction~\cite{wang2024dust3r, wang2025vggt}, which allows for inferring aligned geometry directly from unposed images, a fundamental question arises: \textit{Should geometry remain a mere \textbf{post-hoc ``checker''} for verification, or can it serve as a \textbf{``latent foundation''} that implicitly grounds the entire perception process?}

We introduce the \textbf{Geometry-Grounded Tracking Anything Model (\modelname)}, a unified end-to-end framework for promptable instance tracking in 3D space, taking only unordered images or video as input. At the core of \modelname is the concept of \textbf{\textit{Geometry as Implicit Memory}}: Unlike prior works that rely on explicit temporal memory bank or further post-hoc verification using explicit geometric representations (e.g., depth maps and camera poses), we are the first to unify appearance and geometry within a latent, spatially-aligned feature space to serve as the underlying persistent memory for identity reasoning. To enable seamless support for multi-modal prompting, \modelname embeds text/visual prompts directly into a unified geometric semantic representation through a \textbf{simple yet highly effective cross-modal spatial encoder}. This early-fusion design ensures that recognition and identity reasoning are inherently grounded in the same spatial structure—leading to more precise local recognition and robust cross-view identity stability. Paired with lightweight geometry and mask decoders, \modelname simultaneously performs spatial reconstruction and spatially-consistent instance tracking across large viewpoint changes and complex dynamic scenes—achieving robust 3D-aware tracking without any explicit 3D geometry inputs and temporal memory banks.


Our experimental results demonstrate that the joint training of reconstruction and segmentation significantly enhances cross-view mask consistency, effectively validating geometry as a persistent form of implicit memory. The resulting geometry-aware representation exhibits superior generalization across diverse tasks—including promptable instance spatial tracking, 3D visual grounding, and video object segmentation—maintaining robust performance even under extreme camera motion and dynamic complexity. Notably, \modelname achieves 74.3 S-mIoU on the instance spatial tracking benchmark, outperforming the state-of-the-art SAM2~\cite{ravi2024sam2} (47.6 S-mIoU) by a substantial margin and setting a new milestone for spatially-consistent tracking.

\section{Related Work}
\label{sec:related_work}


\subsection{Geometry Foundation Models}


Traditional 3D reconstruction~\cite{schoenberger2016sfm, hartley2003multiple, pan2024global, schoenberger2016sfm,furukawa2015multi, schonberger2016pixelwise} relies on multi-view geometry for feature matching and pose estimation. Recently, feed-forward models have emerged to directly regress the 3D structure
of a scene from a set of images in a single pass. Dust3R~\cite{wang2024dust3r} predicts point clouds from pairs of images within the reference camera's coordinate system. VGGT~\cite{wang2025vggt} addresses this issue by incorporating multi-task learning and training on large-scale datasets. Pi3~\cite{wang2025pi3} further employs a fully permutation-equivariant architecture to predict affine-invariant camera poses and scale-invariant local point maps without any reference frames. However, these models remain geometry-centric, excelling at spatial reconstruction but lacking the semantic adaptability and interactive capability required in real-world scenarios.



\subsection{Promptable Video Object Segmentation}

Current VOS methods follow two main paradigms: (1) Semi-supervised VOS (Semi-VOS), which tracks objects based on an initial mask~\cite{cheng2022xmem, bekuzarov2023xmem++, yang2021aot, yang2022deaot}, with SAM2 ~\cite{ravi2024sam2} delivering strong generalization and enabling interactive segmentation through a unified prompting framework. (2) Referring VOS (RVOS), which segments objects via language descriptions~\cite{botach2022end,wu2022lreferformer,liang2025referdino}. Despite these advances, current VOS models remain appearance-driven and rely on temporal matching or feature banks, making them sensitive to large camera motions and viewpoint changes. In contrast, our \modelname leverages geometry as implicit memory, enabling consistent object segmentation across challenging views and long-term spatial variations.


\subsection{3D-Aware Instance Tracking}
Recent efforts in egocentric vision have focused on achieving object permanence during drastic camera motion and out-of-view occlusions. 
LMK ~\cite{LMK} introduces the OSNOM task and proposed a ``Lift, Match, and Keep'' framework that projects 2D observations into 3D world coordinates for post-hoc matching. 
Building upon this, the subsequent work~\cite{3dtracking} integrates scene geometry with 2D video object segmentation (VOS) models to refine tracking consistency, which leverages 3D awareness to re-identify objects that have been out of sight for extended periods. While effective, these methods follow a multi-stage ``Tracking-by-Mapping'' paradigm, relying on disconnected off-the-shelf pose and depth estimators which are prone to error propagation. 
In contrast, \modelname avoids such explicit pipelines by unifying appearance and geometry within a latent, spatially-aligned feature space, enabling end-to-end identity reasoning without explicit 3D inputs.

\begin{figure*}[!t]
  \centering
  \includegraphics[width=1\linewidth]{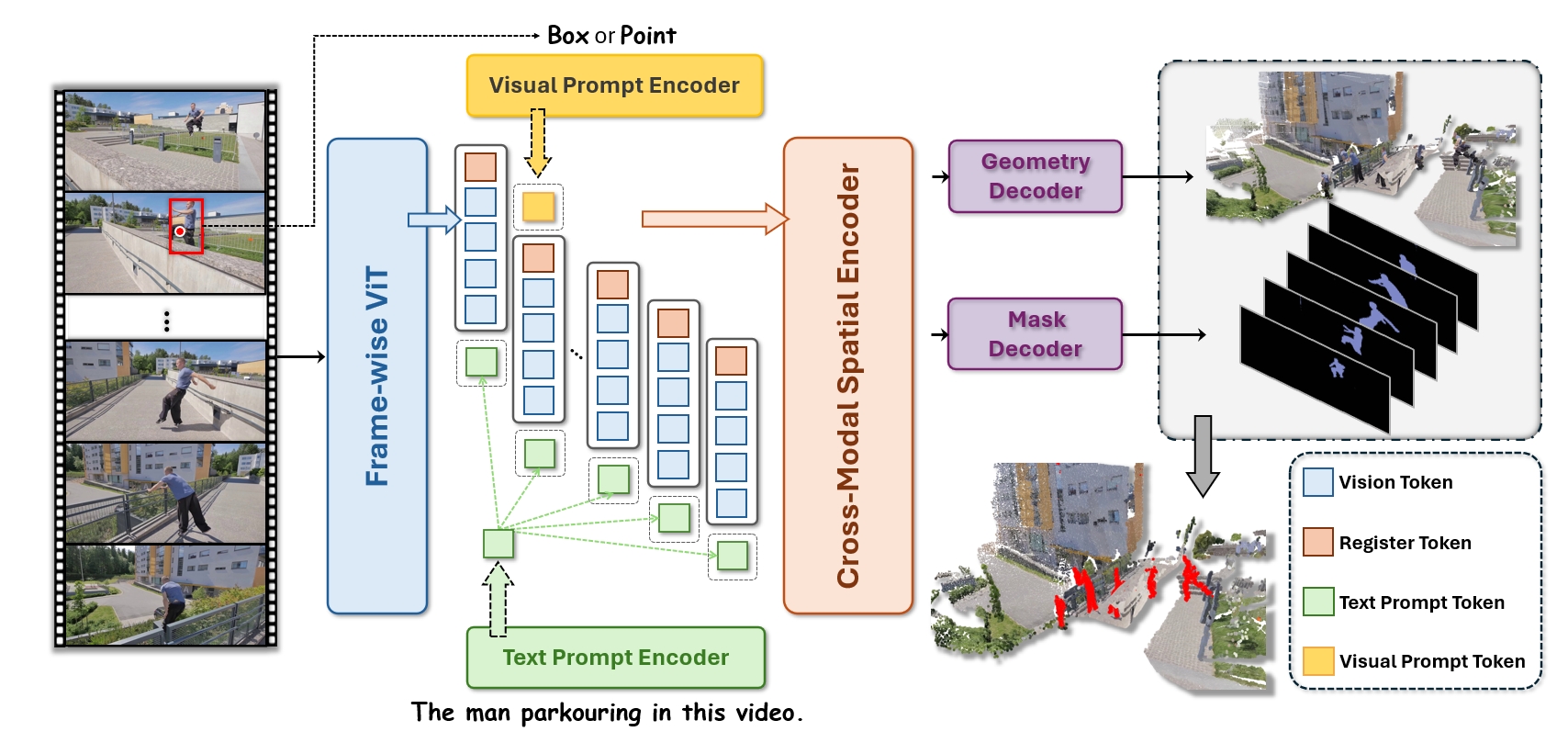}
  \vspace{-2ex}
  \caption{\textbf{Overview of \modelname.} \modelname is a unified framework for spatial reconstruction and promptable instance tracking with 2D prompts in 3D space from unposed RGB observations.}
  \vspace{-2ex}
  \label{fig:model}
\end{figure*}
\section{Method}

In this section, we first outline the architecture of \modelname\ (Sec.~\ref{sec:problem} and~\ref{sec:arch}), followed by the construction pipeline and statistics of the \textit{InsTrack} dataset and benchmark (Sec.~\ref{sec:data}). Finally, we detail the model's training objectives (Sec.~\ref{sec:train}).

\subsection{Problem Formulation}
\label{sec:problem}

The input to our model is a sequence $\mathcal{I} = (\mathbf{I}_i)_{i=1}^N$ of $N$ RGB images $\mathbf{I}_i \in \mathbb{R}^{H \times W \times 3}$ from static or dynamic scenes. The model is conditioned on a prompt $P \in \{P_v, P_t\}$, where $P_v$ represents visual prompts (points or boxes) and $P_t$ represents text prompts. Our \modelname is a function $\Phi$ that maps the image sequence and the prompt to a corresponding set of 3D geometry and consistent instance masks:
\begin{equation}
    \Phi((\mathbf{I}_i)_{i=1}^N, P) = (M_i, G_i)_{i=1}^N
\end{equation}
where for each frame $i$, $M_i \in \{0, 1\}^{H \times W}$ is the instance segmentation mask, and $G_i$ is a comprehensive \textbf{geometry set} defined as:
\begin{equation}
    G_i = \{ \mathbf{K}_i, \mathbf{X}_i, \mathbf{C}_i \}
\end{equation}
Here, $\mathbf{K}_i \in SE(3) \subset \mathbb{R}^{4 \times 4}$ denotes the camera pose, $\mathbf{X}_i \in \mathbb{R}^{H \times W \times 3}$ represents the pixel-aligned 3D point map in the local camera coordinate system, and $\mathbf{C}_i \in [0, 1]^{H \times W}$ is the confidence map providing per-pixel reliability scores for the predicted geometry $\mathbf{X}_i$.

\subsection{Architecture of \modelname}
\label{sec:arch}

As illustrated in Fig.~\ref{fig:model}, our \modelname builds upon Pi3~\cite{wang2025pi3}, and consists of three parts: 1) \textit{Prompt Encoders} to accept various types of visual or text prompts, 2) \textit{Cross-modal Spatial Encoder} to fuse the vision tokens and prompt tokens to construct a unified geometry-semantic representation, and 3) \textit{Geometry \& Mask Decoder} to predict the geometry of the scene and corresponding spatial-temporal consistent object masks for each frame, respectively.

\paragraph{Prompt Encoding.} To facilitate flexible, multi-modal guidance, \modelname projects diverse prompt types into a unified geometric-semantic latent space. Each prompt is encoded into the prompt tokens $\mathbf{T}_p \in \mathbb{R}^{N_p \times D}$, where $N_p$ varies according to the specific prompt modality: 

1) \textit{Visual Prompt Encoder:} Following SAM~\cite{ma2024segment}, we represent visual prompts through a combination of spatial coordinates and semantic identifiers. Specifically, a visual prompt $P_v$ at coordinate $(x, y)$ is transformed into a $D$-dimensional token $\mathbf{t}$ via:
    \begin{equation}
        \mathbf{t} = \text{PE}(x, y) + \mathbf{e}_{type}
    \end{equation}
    where $\text{PE}(\cdot)$ denotes a positional encoding based on random Fourier features, and $\mathbf{e}_{type} \in \mathbb{R}^{D}$ is a learnable embedding distinct to each prompt type (e.g., positive/negative points or box corners). Consequently, a \textit{point prompt} yields $N_p = 1$ token, while a \textit{box prompt} is decomposed into its top-left and bottom-right corners, generating $N_p = 2$ tokens.

2) \textit{Text Prompt Encoder:} Text prompts are tokenized with a pretrained tokenizer and processed through the CLIP~\cite{radford2021clip} text encoder to obtain sentence-level embeddings, which are further transformed by a learnable projection layer to produce the final $N_p = 1$ text token.

\paragraph{Cross-Modal Spatial Encoder.}
Each image $\mathbf{I}_i$ is first embedded into vision tokens $V_i \in \mathbb{R}^{J \times D}$ via a DINOv2~\cite{oquab2023dinov2} backbone. To enable unified spatial reasoning, we define a composite input sequence $Z_i$ for each frame $i$:
\begin{equation}
    Z_i = [V_i \; ; \; \mathbf{T}_{p,i} \; ; \; R]
\end{equation}
where $R$ denotes learnable register tokens. For \textbf{visual prompts}, which are inherently frame-specific, we adopt a localized conditioning strategy: if a frame $i$ contains visual prompts, $\mathbf{T}_{p,i}$ is populated with the corresponding visual prompt tokens; otherwise, it is filled by the \textit{learnable null-prompt embeddings} to maintain structural consistency. In contrast, for \textbf{text prompts}, which provide global semantic context, we adopt a consistent fusion strategy: the text tokens derived from the CLIP encoder are inserted as $\mathbf{T}_{p,i}$ into the sequences of every frame. The resulting sequences $\mathcal{Z} = (Z_i)_{i=1}^N$ are processed through alternating blocks of intra-view self-attention and global cross-view attention to obtain the fused geometric-semantic representation.

\paragraph{Geometry \& Mask Decoder.} 
The fused geometry-semantic representation from each frame is passed to two specialized decoders to produce the final predictions: 1) \textbf{Geometry Decoder}: Inherited from Pi3~\cite{wang2025pi3}, this decoder predicts a comprehensive geometry set $G_i $ for each frame. 2) \textbf{Mask Decoder}: Adapted from SAM~\cite{kirillov2023segment}, this decoder predicts the instance mask $M_i$. Due to the early multi-modal fusion mechanism introduced by our cross-modal spatial encoder, we eliminate the prompt embeddings during mask decoding. Besides, we append an extra token alongside the mask and IoU tokens, and apply an additional MLP head to this token to predict the likelihood that the target object is present in the frame.

\subsection{InsTrack Dataset and Benchmark}
\label{sec:data}

\paragraph{Data Collection Pipeline.}To facilitate promptable instance tracking, we develop an automated annotation pipeline to generate multi-modal prompts and instance-consistent masks for the ScanNet++ dataset~\cite{yeshwanth2023scannet++}. The pipeline streamlines three stages: (1) \textit{Image Subsampling}: We employ pose-based filtering~\cite{zhang2025SPAR} to remove $\sim$80\% of redundant frames while maintaining maximal scene coverage. (2) \textit{Mask Generation}: 2D instance masks are obtained by rasterizing 3D mesh annotations and establishing pixel-to-face mappings. (3) \textit{Multi-modal Prompting}: We randomly sample up to 20 objects per scene to generate \textit{visual prompts} ( points and boxes) and integrate existing human-annotated or reasoning-based \textit{text prompts} from L3DD~\cite{arnaud2025locate3d} and SURPRISE3D~\cite{huang2025surprise3d}. We provide more data collection details in Appendix~\ref{appendix:pipeline}.

\paragraph{InsTrack Dataset.} Following the official ScanNet++ splits, the \textbf{InsTrack} dataset comprises 856 training and 50 validation scenes. Each sample provides RGB images, multi-modal prompts, depth maps, camera poses, and 3D-consistent instance masks. In total, the training set encompasses 99,666 text and 77,007 visual prompts (64,213 points, 12,794 boxes), while the validation set contains 3,784 text and 2,018 visual prompts (1,270 points, 748 boxes). We partition the benchmark into \textit{InsTrack-Text} and \textit{InsTrack-Visual} to individually assess the model's proficiency across different prompt modalities.

\subsection{Model Training}
\label{sec:train}

To achieve promptable instance tracking capabilities while maintaining the original geometry reconstruction capabilities, our model is trained end-to-end by minimizing a composite loss function $\mathcal L$, which is a weighted sum of the segmentation loss $\mathcal{L}_{\text{seg}}$, and the geometry loss $\mathcal{L}_{\text{geo}}$, which consists of the point reconstruction loss, the confidence loss, and the camera pose loss:

\begin{equation}
\label{eq:total_loss_new}
\mathcal{L}=\lambda_{\text{seg}}\mathcal{L}_{\text{seg}}+\lambda_{\text{geo}}\mathcal{L}_{\text{geo}} .
\end{equation}
\begin{equation}
\label{eq:geometry_loss}
\mathcal{L}_{\text{geo}}=\mathcal{L}_{\text{points}}+\lambda_{\text{normal}}\mathcal{L}_{\text{normal}}+\lambda_{\text{conf}}\mathcal{L}_{\text{conf}}+\lambda_{\text{cam}}\mathcal{L}_{\text{cam}} .
\end{equation}

To ensure generalization and robustness, we train the model on a large-scale aggregation of datasets, including segmentation datasets, reconstruction datasets, and joint segmentation-reconstruction datasets. Details of model training can be found in Appendix~\ref{appendix:training_details}.

\begin{figure}[t]
  \centering
  \includegraphics[width=1\linewidth]{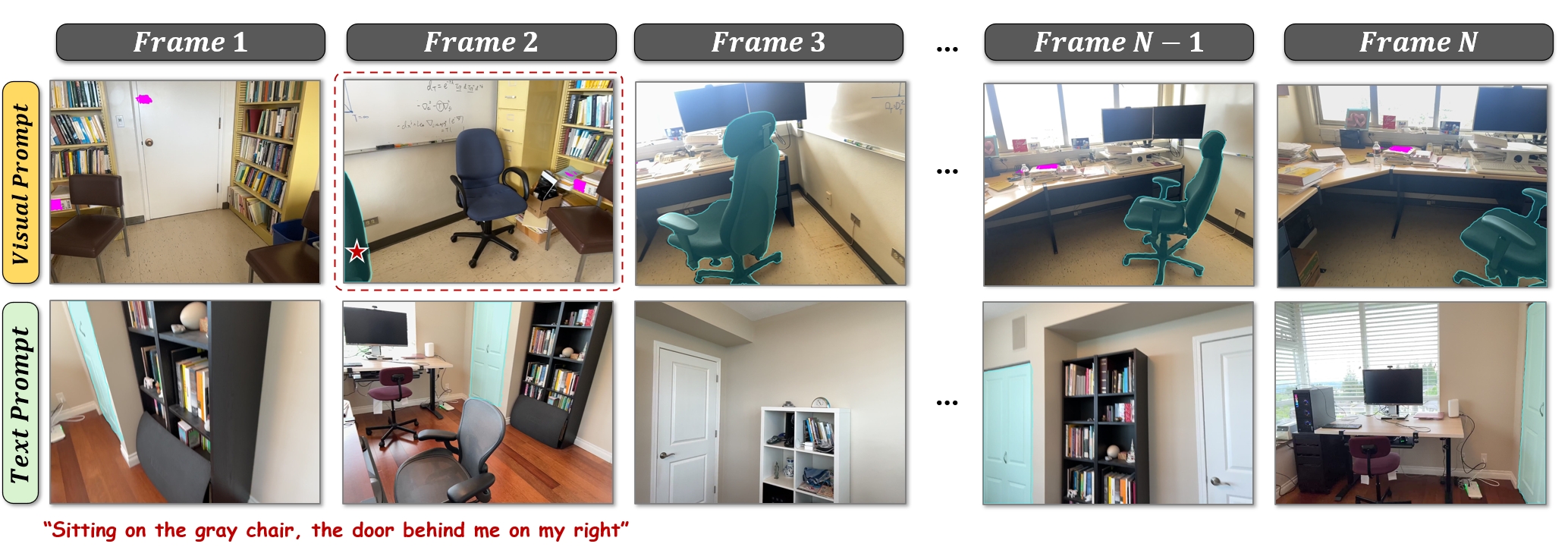}
  \caption{\textbf{Qualitative results on InsTrack validation set. } The top and bottom rows illustrate that our \modelname{} effectively handles diverse prompt types and maintains spatially consistent instance tracking across frames.}
  \label{fig:vis_results}
\end{figure}

\section{Experiment}

\subsection{Promptable Instance Spatial Tracking}

In this section, we introduce a new task: Promptable Instance Spatial Tracking (PIST), designed to evaluate a model's instance segmentation spatial consistency across views in the static scenario. Given an input prompt—such as a point, bounding box on any view, or a referring text expression—the goal is to produce spatially consistent masks for the corresponding instance across all views.
For the PIST task, we evaluate performance using Spatial mIoU (S-mIoU) and Spatial Success Rate (S-SR).  
Given an object $o$ and its predicted masks $\{\hat{M}_n^o\}_{n=1}^N$ across $N$ views with ground-truth masks $\{M_n^o\}_{n=1}^N$, S-mIoU is defined as

\begin{equation}
\label{eq:s-miou}
\text{S-mIoU}(o) = \frac{1}{N} \sum_{t=1}^{N} 
\frac{|\hat{M}_n^o \cap M_n^o|}{|\hat{M}_n^o \cup M_n^o|}
\end{equation}

S-SR measures whether the object is consistently tracked across all views. A frame is considered successfully tracked only if the frame IoU exceeds 0.5.  
Formally,

\begin{equation}
\label{eq:s-sr}
\text{S-SR}(o) = \mathbb{1} \left[ \forall n \in \{1, \ldots, N\}, \frac{|\hat{M}_n^o \cap M_n^o|}{|\hat{M}_n^o \cup M_n^o|} > 0.5 \right]
\end{equation}

We benchmark \modelname against strong video object segmentation methods on the InsTrack validation set (Tab.~\ref{tab:instrack}).
For InsTrack-\textit{Visual}, we compare to Cutie~\cite{cheng2024putting} and SAM2~\cite{ravi2024sam2}, while for InsTrack-\textit{Text}, we include the recent referring video object segmentation models ReferFormer~\cite{wu2022lreferformer} and ReferDINO~\cite{liang2025referdino} to ensure a fair comparison. Across all settings, our \modelname establishes a clear performance margin. On the \textit{Text} part, our method achieves 72.3 S-mIoU and 77.6 S-SR, substantially outperforming ReferFormer (37.6 / 43.7) and ReferDINO (41.7 / 48.2). On the \textit{Visual} part, \modelname further improves performance to 75.8 S-mIoU and 81.2 S-SR, surpassing SAM2 and Cutie-base by large margins.
Aggregated over both subsets, our model attains 74.3 S-mIoU and 80.1 S-SR, marking a significant improvement over all baselines. These results highlight the benefit of using geometry-aware representations as implicit memory.
Unlike conventional video segmentation models—which tend to lose track under substantial viewpoint change or extended temporal spans, often yielding S-mIoU scores below 50—\modelname reliably maintains object identity across views and achieves strong spatial tracking success, with the qualitative results in Fig.~\ref{fig:vis_results}.


\begin{table*}[t]
\centering
\setlength{\tabcolsep}{2pt}
\resizebox{0.95\textwidth}{!}{
\arrayrulecolor{black}
\begin{tabular}{lcccccccc}
\toprule
\multicolumn{1}{c}{\textbf{Model}} & \multicolumn{3}{c}{\textbf{Capability}} 
& \multicolumn{3}{c}{\textbf{Spatial Track (S-mIoU / S-SR)}} 
& \multicolumn{2}{c}{\textbf{Recon. Metric}}  \\
\cmidrule(lr){2-4} \cmidrule(lr){5-7} \cmidrule(lr){8-9}
& Recon. & Understand & Track 
& \textit{Text} & \textit{Visual} & \textit{Overall}
& Abs. Rel$\downarrow$ & $\tau\uparrow$ \\
\midrule
ReferFormer~\cite{wu2022lreferformer} & \xmark & \cmark & \cmark & 37.6 / 43.7 & - / - & 37.6 / 43.7 & - & - \\
ReferDINO~\cite{liang2025referdino} & \xmark & \cmark & \cmark & 41.7 / 48.2 & - / - & 41.7 / 48.2 & - & - \\
Cutie-base~\cite{cheng2024putting} & \xmark & \cmark & \cmark & - / - & 42.7 / 51.9 & 42.7 / 51.9 & - & - \\
SAM2~\cite{ravi2024sam2} & \xmark & \cmark & \cmark & - / - & 47.6 / 53.1 & 47.6 / 53.1  & - & - \\
\midrule
VGGT~\cite{wang2025vggt} & \cmark & \xmark & \xmark & - & - & - & 2.67 & 85.87 \\
Pi3~\cite{wang2025pi3} & \cmark & \xmark & \xmark & - & - & - & 2.54 & 86.72 \\
\rowcolor{myblue!25}
\midrule
\textbf{\modelname (Ours)} & \cmark & \cmark & \cmark & \textbf{72.3 / 77.6} & \textbf{75.8 / 81.2} & \textbf{74.3 / 80.1} & \textbf{2.51} & \textbf{86.91} \\
\bottomrule
\end{tabular}
}
\vspace{4pt}
\caption{\textbf{Quantitative results on InsTrack validation set.} We report the instance spatial tracking quality on the \textit{Text} part, \textit{Visual} part, and \textit{Overall}, together with reconstruction accuracy metrics. \textbf{Bold} indicates the best results.}
\label{tab:instrack}
\end{table*}


\subsection{3D Visual Grounding}

3D Visual Grounding aims to localize a target object in 3D scene given a natural-language referring expression. In this section, we study the performance of \modelname on three standard visual grounding benchmarks: SR3D~\cite{ReferIt3D}, NR3D~\cite{ReferIt3D}, and ScanRefer~\cite{ScanRefer}. To unify the evaluation, we follow the previous method~\cite{arnaud2025locate3d} to evaluate all the methods by reporting the top-1 accuracy without assuming ground-truth object bounding boxes under Acc@0.25 and Acc@0.5 metrics. \modelname can directly reconstruct 3D scenes and predict instance-consistent masks across images. However, the reconstructed point clouds output by our model lack absolute scale information, making it difficult to perform a fair comparison in 3D space. To ensure accurate quantitative comparison, we therefore employ ground-truth depth and camera poses to project each predicted mask across views into 3D space to obtain 3D instance point clouds masks. In addition, to improve the quality of the final obtained 3D bounding boxes, we apply a post-processing stage that filters out noise points to determine the final 3D bounding box.

As shown in Tab.~\ref{tab:scanrefer_vg}, our \modelname significantly outperforms all previous zero-shot approaches on both NR3D and ScanRefer. In particular, it surpasses the prior state-of-the-art method, VLM-Grounder, by a large margin—achieving 45.7\% overall Acc@0.5 compared to VLM-Grounder’s 32.8\% on ScanRefer. Notably, VLM-Grounder relies on a combination of a VLM with external detection and segmentation tools to obtain 2D object masks across views, which are then projected and fused into the 3D point cloud. This multi-stage process leads to a large performance gap (18.8) between Acc@0.25 and Acc@0.5. In contrast, our \modelname inherently maintains geometric consistency, yielding far more stable view alignment and a smaller gap (10.5) between Acc@0.25 and Acc@0.5. Moreover, even without using explicit 3D geometry from point clouds as inputs, \modelname still outperforms the previous point-cloud-based methods, 3D-VisTA, which were trained and evaluated on the same dataset, demonstrating its strong capability to infer object relationships directly from 2D images.

To verify that the visual grounding results do not depend on ground-truth geometry at inference time, we further evaluate 3D projection using the depth and camera pose predicted by \modelname. As shown in Tab.~\ref{tab:predicted_geometry_vg}, replacing ground-truth geometry with predicted geometry leads to only a small drop on ScanRefer and SR3D. This confirms that \modelname learns sufficiently accurate RGB-only geometry to support high-quality 3D instance grounding.

\begin{table*}[t!]
\centering
\scalebox{0.95}{\tablestyle{2.2pt}{1.15}

\begin{tabular}{lccccccccc}
\toprule
                                    \multirow{2}{*}{Method}    & &  &   &  \multicolumn{2}{c}{SR3D} & \multicolumn{2}{c}{NR3D} & \multicolumn{2}{c}{ScanRefer} \\ \cmidrule(l){5-6} \cmidrule(l){7-8} \cmidrule(l){9-10} 
                             & w/o. LLM/VLM & w/o. PC  & Zero-Shot & Acc@0.25   & Acc@0.5        & Acc@0.25  & Acc@0.5        & Acc@0.25      & Acc@0.5      \\ \midrule
ScanRefer~\cite{ScanRefer}              & \cmark    & \xmark  & \xmark  & -      & -           & -      & -           & 35.5          & 22.4         \\
LanguageRefer~\cite{Languagerefer}       & \cmark    & \xmark  & \xmark & 39.5       & -           & 28.6    & -          & -          & -         \\
InstanceRefer~\cite{instancerefer}      & \cmark    & \xmark  & \xmark & 31.5       & -          & 29.9     & -           & 40.2        & 32.9         \\
SAT-2D~\cite{instancerefer}      & \cmark    & \xmark  & \xmark & 35.4       & -          & 31.7    & -           & 44.5        & 30.1        \\
3D-VisTA~\cite{3d-vista} & \cmark    & \xmark  & \xmark & 56.5       & 51.5       & 47.7      & 42.2           & 51.0          & 46.2         \\
BUTD-DETR~\cite{butd-detr}              & \cmark    & \xmark  & \xmark  & 52.1      & -          & 43.3      & -           & 52.2          & 39.8          \\ \midrule
OpenScene~\cite{OpenScene}              & \cmark    & \xmark   & \cmark & -       & -            & -     & -          & 13.2          & 6.5         \\
LLM-Grounder~\cite{LLM-Grounder}        & \xmark    & \xmark  & \cmark & -       & -          & -         & -              & 17.1            & 5.3           \\
VLM-Grounder~\cite{vlm-grounder}  & \xmark & \cmark & \cmark  & - & -   & 48.0 & -          & 51.6 & 32.8 \\ \midrule
\rowcolor{myblue!25}
\textbf{\modelname (Ours)} & \cmark & \cmark  & \cmark &  \textbf{57.2} & \textbf{48.7} & \textbf{51.4} & \textbf{45.8} & \textbf{56.2} & \textbf{45.7} \\ \bottomrule
\end{tabular}
}
\vspace{4pt}
\caption{\textbf{3D visual grounding results on SR3D, NR3D, and ScanRefer.} We evaluate top-1 accuracy on the validation set without any assumption of ground-truth proposals. \modelname outperforms previous LLM/VLM-assisted agent methods and point-cloud input-based methods in a zero-shot manner.}
\label{tab:scanrefer_vg}
\end{table*}

\begin{table}[t]
\centering
\resizebox{\columnwidth}{!}{
\begin{tabular}{lcc}
\toprule
\textbf{3D Projection Source} & \textbf{SR3D Acc@0.5} & \textbf{ScanRefer Acc@0.5} \\
\midrule
GT Depth + GT Pose & 48.7 & 45.7 \\
\rowcolor{myblue!25}
Pred. Depth + Pred. Pose & 47.3 & 44.9 \\
\bottomrule
\end{tabular}
}
\vspace{4pt}
\caption{\textbf{3D visual grounding with predicted geometry.} \modelname remains close to the evaluation setting using ground-truth depth and pose, showing that it is not dependent on ground-truth geometry at inference time.}
\label{tab:predicted_geometry_vg}
\end{table}

\begin{figure*}[t]
  \centering
  \includegraphics[width=1\linewidth]{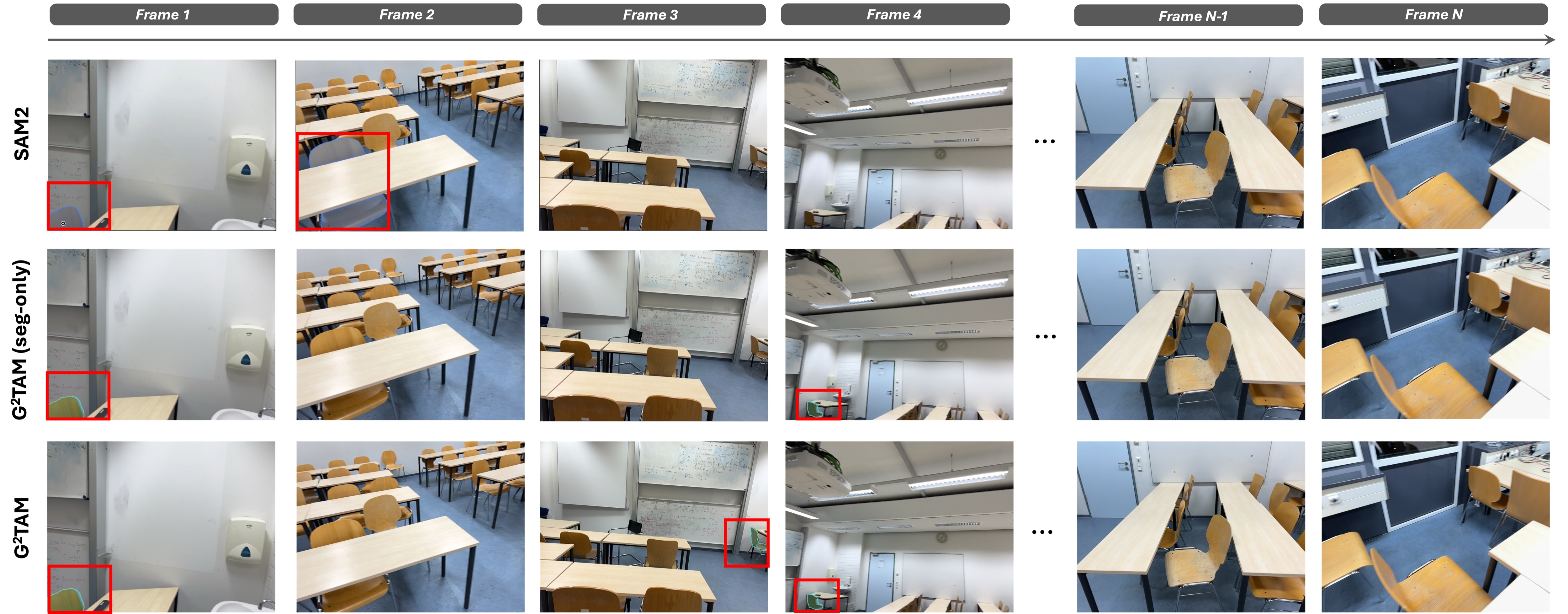}
  \caption{\textbf{Qualitative comparison of \modelname variants against SAM2.} By joint training on reconstruction and segmentation, \modelname achieves robust spatio-temporal consistency, maintaining precise tracking even in challenging sequences where baselines fail.}
  \label{fig:compare}
\end{figure*}

\subsection{Promptable Video Object Segmentation}

Since no existing video object segmentation benchmarks support both text and visual prompts, to evaluate our model’s instance tracking performance in dynamic scenarios, we conduct experiments under two settings: (1) \textit{Semi-supervised Video Object Segmentation} (Semi-supervised VOS), where the prompt is a ground-truth mask provided on the first frame, and (2) \textit{Referring Video Object Segmentation} (RVOS), where the prompt is a natural-language referring expression. Unlike Promptable Instance Spatial Tracking, which primarily evaluates spatial consistency in static scenes, both Semi-supervised VOS and RVOS focus on dynamic scenarios where target objects undergo motion over time. These two settings require the model to preserve spatial accuracy and temporal coherence in segmentation predictions under complex motion and appearance variations, providing a comprehensive assessment of instance consistency in dynamic environments. For video object segmentation tasks, we report the performance using standard protocols J\&F~\cite{pont20172017}. 

We compare \modelname with prior approaches in Tab.~\ref{tab:semi-vos} for Semi-supervised VOS and Tab.~\ref{tab:rvos} for RVOS. As shown in Tab.~\ref{tab:semi-vos}, \modelname consistently outperforms SAM2~\cite{ravi2024sam2}. To ensure a rigorous and fair comparison, we evaluate both models at a consistent inference resolution of $512 \times 512$. Notably, our model achieves the most significant performance gain on the MOSE benchmark---specifically designed to assess tracking under heavy occlusion---where our score improves from $75.2$ to $\mathbf{77.8}$. This substantial margin underscores the efficacy of \textbf{geometric grounding}, which empowers \modelname to maintain robust object correspondence even when visual cues are severely degraded. On RVOS benchmarks (Tab.~\ref{tab:rvos}), \modelname achieves 72.2 \mjf on Ref-YTVOS and 71.7 on Ref-DAVIS17, significantly outperforming the previous SOTA ReferDINO~\cite{liang2025referdino} by +2.9 and +2.8, respectively.
On the more challenging MeViS benchmark~\cite{ding2023mevis}, which focuses on referring motion expressions, \modelname also improves over ReferDINO from 49.3/44.7/53.9 to 51.2/46.3/55.7 in \mjf/\mj/\mf.
Notably, \modelname performs these tasks without maintaining the explicit memory bank, demonstrating that the proposed global geometry representation inherently preserves instance correspondence across time, serving as a unified foundation for both static and dynamic visual understanding.

\begin{table*}[t]
\centering
\begin{small}
\begin{tabular}{lccccc}
\toprule
\multirow{2}{*}{Method} & \multicolumn{4}{c}{\mjf} & \mg \\ 
\cmidrule(lr){2-5}\cmidrule(lr){6-6}
 & MOSE val & DAVIS 2017 val  & SA-V val & SA-V test & YTVOS 2019 val \\ 
\midrule
STCN~\cite{cheng2021rethinking} & 52.5 & 85.4  & 61.0 & 62.5 & 82.7 \\
SwinB-AOT~\cite{yang2021aot} & 59.4 & 85.4 & 51.1 & 50.3 & 84.5 \\
SwinB-DeAOT~\cite{yang2022deaot} & 59.9 & 86.2  & 61.4 & 61.8 & 86.1 \\
RDE~\cite{li2022recurrent} & 46.8 & 84.2 & 51.8 & 53.9 & 81.9 \\
XMem~\cite{cheng2022xmem} & 59.6 & 86.0  & 60.1 & 62.3 & 85.6 \\
SimVOS-B~\cite{wu2023scalable} & - & 88.0  & 44.2 & 44.1 & 84.2 \\
JointFormer~\cite{zhang2023joint} & - & 90.1  & - & - & 87.4 \\
ISVOS~\cite{wang2023look} & - & 88.2  & - & - & 86.3 \\
DEVA~\cite{cheng2023deva} & 66.0 & 87.0 & 55.4 & 56.2 & 85.4 \\
Cutie-base~\cite{cheng2024putting} & 69.9 & 87.9  & 60.7 & 62.7 & 87.0 \\
Cutie-base+~\cite{cheng2024putting} & \underline{71.7} & 88.1  & 61.3 & 62.8 & 87.5 \\
\midrule
SAM 2 ~\cite{ravi2024sam2}  &
75.2 & 89.4 & 75.8 & 76.7 & 87.8 \\
\rowcolor{myblue!25}
\textbf{\modelname (Ours)} & \textbf{77.8} & \textbf{89.9} & \textbf{76.8} & \textbf{77.6} & \textbf{89.1} \\
\bottomrule
\end{tabular}
\end{small}
\vspace{4pt}
\caption{\textbf{Semi-supervised VOS results on various benchmarks.} \modelname achieves comparable performance with powerful SAM2 in accuracy (\mjf, \mg) for video segmentation based on first-frame ground-truth mask prompts.
}
\label{tab:semi-vos}
\end{table*}

\begin{table}[t]
\centering
\begin{small}
\resizebox{\columnwidth}{!}{
\begin{tabular}{lcccccc}
\toprule
\multirow{2}{*}{Method} & \multicolumn{3}{c}{Ref-YouTube-VOS} & \multicolumn{3}{c}{Ref-DAVIS17} \\ 
\cmidrule(lr){2-4}\cmidrule(lr){5-7}
 & \mjf & \mj  & \mf & \mjf & \mj & \mf \\ 
\midrule
ReferFormer~\citep{wu2022lreferformer} & 62.9 & 61.3 &64.6 & 61.1 & 58.1 &64.1 \\
HTML~\citep{han2023html} & 63.4  & 61.5 &  65.2  &  62.1 &  59.2  & 65.1 \\
SgMg~\citep{miao2023sgmg} & 65.7 &63.9 &67.4 &63.3 &60.6 &66.0 \\
ReferDINO~\citep{liang2025referdino} & 69.3 &67.0 &71.5 &68.9 &65.1 &72.9 \\
\midrule
\rowcolor{myblue!25}
\textbf{\modelname (Ours)} & \textbf{72.2} & \textbf{69.1}  & \textbf{73.1} & \textbf{71.7} & \textbf{68.2}  & \textbf{75.1}\\
\bottomrule
\end{tabular}}
\end{small}
\caption{\textbf{RVOS results on various benchmarks.}
}
\label{tab:rvos}
\end{table}

\subsection{Inference Efficiency}

We report matched inference statistics on a single A100 GPU using the same setting for SAM2 and \modelname ($512 \times 512$ input resolution with 8 frames). As shown in Tab.~\ref{tab:efficiency}, \modelname introduces moderate overhead compared with SAM2, but this cost provides additional capabilities including RGB-only geometry prediction, text-conditioned spatial tracking, and 3D visual grounding. Profiling in Tab.~\ref{tab:runtime_breakdown} shows that the main cost comes from frame-wise attention and global cross-view attention, which are responsible for geometry-aware multi-view reasoning.

\begin{table}[t]
\centering
\resizebox{\columnwidth}{!}{
\begin{tabular}{lccc}
\toprule
\textbf{Method} & \textbf{Peak Mem. (GB)} & \textbf{FPS} & \textbf{Time (s)} \\
\midrule
SAM2~\cite{ravi2024sam2} & 3 & \textbf{30.2} & \textbf{0.26} \\
\rowcolor{myblue!25}
\textbf{\modelname (Ours)} & 4 & 21.6 & 0.37 \\
\bottomrule
\end{tabular}
}
\vspace{4pt}
\caption{\textbf{Inference efficiency comparison with SAM2.} Both methods are evaluated on a single A100 GPU at $512 \times 512$ resolution with 8 input frames.}
\label{tab:efficiency}
\end{table}

\begin{table*}[t]
\centering
\resizebox{\textwidth}{!}{
\begin{tabular}{lccccc}
\toprule
\textbf{Component} & Encoder & Frame-wise attention & Global cross-view attention & Mask head & Reconstruction head \\
\midrule
\textbf{Time (s)} & 0.0720 & 0.0965 & 0.1395 & 0.0610 & 0.0521 \\
\bottomrule
\end{tabular}
}
\vspace{4pt}
\caption{\textbf{Runtime breakdown of \modelname.} The attention modules dominate inference time, indicating clear future directions for sparse or more efficient cross-view interaction.}
\label{tab:runtime_breakdown}
\end{table*}

\subsection{3D Scene Reconstruction}

For reconstruction evaluation, we follow Pi3~\cite{wang2025pi3} to adopt Absolute Relative Error (Abs.\ Rel) and the threshold accuracy metric ($\delta{<}1.03$) for scene-level assessment.
As shown in Tab.~\ref{tab:instrack}, our method attains lower Abs.\ Rel and higher $\delta$ accuracy than both Pi3 and VGGT on the InsTrack validation set.
To further examine generalization, we evaluate scale-invariant monocular depth on the additional datasets using the standard Abs.\ Rel and $\delta{<}1.25$ metrics. Tab.~\ref{tab:monodepth} demonstrates that our approach achieves superior monocular depth accuracy compared to the Pi3.

\begin{table}[t]
    \centering
    \vspace{-1em}
    \tablestyle{1pt}{1.05}
    \resizebox{1.0\columnwidth}!{
    \begin{tabular}{lcccccc}
        \toprule[0.17em]
        \multirow{3}{*}{\textbf{Method}} &
        \multicolumn{2}{c}{\textbf{Sintel}} &
        \multicolumn{2}{c}{\textbf{KITTI}} &
        \multicolumn{2}{c}{\textbf{NYU-v2}} \\
        \cmidrule(r){2-3} \cmidrule(r){4-5} \cmidrule(r){6-7} 
        &
        Abs Rel$\downarrow$ & $\delta<1.25\uparrow$ &
        Abs Rel$\downarrow$ & $\delta<1.25\uparrow$ &
        Abs Rel$\downarrow$ & $\delta<1.25\uparrow$ \\
        \midrule%
        DUSt3R~\cite{wang2024dust3r} & 0.488 & 0.532  & 0.109 & 0.873 & 0.081 & 0.909 \\
        MASt3R~\cite{leroy2024grounding} & 0.413 & 0.569 &0.077 &0.948 &0.110 &0.865 \\
        MonST3R~\cite{zhang2024monst3r} & 0.402 & 0.525  &0.098 & 0.895 & 0.094 & 0.887 \\
        Fast3R~\cite{yang2025fast3r} & 0.544 & 0.509  & 0.120 & 0.861 & 0.093 & 0.898 \\
        CUT3R~\cite{wang2025continuous} & 0.418 & 0.520 & 0.097 & 0.914 & 0.081 & 0.914 \\
        FLARE~\cite{zhang2025flare} & 0.606 & 0.402  & 0.312 & 0.513 & 0.089 & 0.898 \\
        VGGT~\cite{wang2025vggt} & 0.335 & 0.599  & 0.082 & 0.947 & 0.056 & 0.951 \\
        Pi3~\cite{wang2025pi3} &\underline{0.277} &\underline{0.614} &\underline{0.060} &\underline{0.971} &\underline{0.054} &\underline{0.956} \\
         \midrule
        \rowcolor{myblue!25}
        \textbf{\modelname (Ours)} & \textbf{0.275} & \textbf{0.616} & \textbf{0.059} & \textbf{0.973} & \textbf{0.052}  & \textbf{0.959} \\
        \bottomrule
    \end{tabular}
    }
    \vspace{4pt}
    \caption{
        \textbf{Monocular Depth Estimation on Sintel~\cite{bozic2021transformerfusion}, KITTI~\cite{geiger2013kitti} and NYU-v2~\cite{silberman2012indoor}.} 
    }
    \vspace{4pt}
    \label{tab:monodepth}

\end{table}

\subsection{Ablation Study}
In this section, we systematically evaluate the core components of \modelname. We first establish a strong baseline, termed \textit{SegPi3}, by simply extending the original Pi3 with a SAM-style mask decoder to support instance segmentation. Based on this, we analyze the impact of our architectural innovations and training strategies.

\paragraph{Architecture Design and Text Integration.} We begin by justifying our architectural choice against the SegPi3 baseline. As shown in Tab.~\ref{tab:ablation_prompt_injection}, SegPi3 treats text embeddings as sparse prompts injected directly into the decoder. However, this late-fusion approach results in a significant performance drop (from 72.3 to 61.8) on the InsTrack \textit{Text} validation set. In contrast, our proposed cross-modal spatial encoder facilitates an early-fusion paradigm, demonstrating that integrating semantic guidance at the encoding stage is crucial for complex multi-modal reasoning. Building upon our early-fusion encoder, we further investigate the optimal strategy for token-level text injection. We experiment with inserting text tokens between vision and register tokens across varying numbers of frames ($1, 2, 5,$ or all). Our results in Tab.~\ref{tab:ablation_prompt_injection} reveal that treating the text prompt as a global conditioning signal—by injecting it across all frames—achieves the best performance. 





\begin{table}[t]
\centering
\resizebox{\columnwidth}{!}{
\begin{tabular}{lccc}
\toprule
\textbf{Method} & \textbf{Frame Num} & \textbf{InsTrack \textit{Text}}& \textbf{Ref-YouTube-VOS} \\
\midrule
SegPi3 (baseline) & all & 61.8 & 63.2\\
\midrule
\multirow{4}{*}{\textbf{\modelname (Ours)}} 
 & 1  & 58.8 & 59.9 \\
 & 2  & 63.2 & 61.7\\
 & 5  & 67.9 & 65.4 \\
 \rowcolor{myblue!25}
 & all & \textbf{72.3} & \textbf{70.2}\\
\bottomrule
\end{tabular}
}
\vspace{4pt}
\caption{Ablation on architecture design and text integration strategies on InsTrack \textit{Text} and Refer-YouTube-VOS benchmarks.}
\label{tab:ablation_prompt_injection}
\end{table}

\begin{table}[t]
\centering
\resizebox{\columnwidth}{!}{
\begin{tabular}{lcccc}
\toprule
\textbf{Method}  & \textbf{InsTrack \textit{Visual}} & \textbf{SA-V val} & \textbf{MOSE} & \textbf{DAVIS} \\
\midrule
 \rowcolor{myblue!25}
\modelname & \textbf{75.8} & 73.9 & 77.8 & \textbf{89.9} \\
\modelname + Memory Module  & 74.9 & \textbf{74.2} & \textbf{78.2} & 89.5 \\ 
\bottomrule
\end{tabular}
}
\vspace{4pt}
\caption{Ablation on explicit memory modules. All VOS columns report \mjf.}
\label{tab:ablation_explicit_memory}
\end{table}

\begin{table}[t]
\centering
\resizebox{1\columnwidth}{!}{
\begin{tabular}{lccc}
\toprule
\textbf{Method} & \textbf{Training Data}  & \textbf{InsTrack \textit{Text}} & \textbf{InsTrack \textit{Visual}} \\
\midrule
\modelname(seg-only) & Seg & 68.2 & 72.7 \\
\rowcolor{myblue!25}
\modelname & Seg + Recon   & \textbf{72.3} & \textbf{75.8} \\ 
\bottomrule
\end{tabular}
}
\vspace{4pt}
\caption{Ablation on joint reconstruction and segmentation training on InsTrack validation datasets.}
\label{tab:ablation_seg_recon}
\end{table}

\paragraph{Does Explicit Memory Help?}

In this section, we investigate whether integrating a conventional explicit memory bank (as used in SAM2) could further bolster our model. We provide the full architectural details of this integration in Appendix~\ref{appendix:combine}. Results in Tab.~\ref{tab:ablation_explicit_memory} show that explicit memory is not uniformly beneficial: it slightly improves SA-V and MOSE, but degrades InsTrack \textit{Visual} and DAVIS. This suggests that geometry-aligned implicit memory already captures most of the benefit, while combining it with explicit memory requires more careful design.


\paragraph{How Geometry Benefits Segmentation?}


We investigate whether the reconstruction objective truly bolsters spatio-temporal consistency in segmentation. First, replacing our Pi3-based encoder with a SAM2 encoder results in a significant performance drop (from 61.8 to 54.2) on InsTrack \textit{Text} part, proving that geometry pretraining yields superior spatial alignment. Furthermore, we compare our joint-training paradigm against a segmentation-only baseline, \modelname (seg-only) in Tab.~\ref{tab:ablation_seg_recon}. Notably, joint training yields a substantial 4.0\% improvement in segmentation accuracy. 
Besides, we provide a qualitative comparison between \modelname, \modelname (seg-only), and SAM2 to evaluate their tracking and segmentation capabilities in Fig.~\ref{fig:compare} . Given a visual prompt for a specific chair in Frame 1, SAM2 incorrectly segments a different chair in Frame 2 and fails to detect the target object in Frames 3 and 4. In contrast, \modelname (seg-only), which is trained exclusively on segmentation data, successfully identifies the target chair in Frame 4 but fails to accurately recognize it in Frame 3. However, only the full \modelname, leveraging the reconstruction objective as a geometric regularizer, maintains seamless tracking across all frames (e.g., Frames 3 and 4). This demonstrates that joint training effectively stabilizes the model's spatial memory, ensuring that promptable segmentation remains coherent and robust across diverse viewpoints.

\section{Conclusion}

In this paper, we presented \modelname, a unified framework for promptable instance tracking in 3D space, given only unordered images or video as input. The key to our success is that we propose that spatially aligned geometric representations can serve as the implicit memory to achieve instance identity and localization across views and time in
both static and dynamic scenes. To achieve this, we present a new promptable instance tracking dataset and benchmark, InsTrack, and the Promptable Instance Spatial Tracking (PIST) task. By joint training on segmentation and reconstruction datasets, our \modelname could achieve robust instance tracking under large viewpoint variations and long-term sequences, facilitating a wide range of interactive and geometry-grounded applications.

\section*{Acknowledgements}
This work is partially supported by the National Natural Science Foundation of China (No. 62402406).

\section*{Impact Statement}
This work contributes to visual-spatial intelligence. By improving the spatial reasoning and temporal tracking capabilities of AI systems, our research has potential applications in areas such as autonomous driving, robotic manipulation, and augmented reality. While these advancements offer significant societal benefits in automation and safety, we acknowledge the general ethical considerations associated with computer vision, including privacy concerns and the potential for misuse in surveillance. However, as this study focuses on fundamental algorithmic improvements using publicly available datasets, we do not foresee any immediate specific negative societal impacts.

\nocite{langley00}

\bibliography{example_paper}

@inproceedings{langley00,
 author    = {P. Langley},
 title     = {Crafting Papers on Machine Learning},
 year      = {2000},
 pages     = {1207--1216},
 editor    = {Pat Langley},
 booktitle     = {Proceedings of the 17th International Conference
              on Machine Learning (ICML 2000)},
 address   = {Stanford, CA},
 publisher = {Morgan Kaufmann}
}

@String(CVPR= {IEEE Conf. Comput. Vis. Pattern Recog.})

@String(ICCV= {Int. Conf. Comput. Vis.})

@String(ECCV= {Eur. Conf. Comput. Vis.})

@String(CVPR  = {CVPR})

@String(ICCV  = {ICCV})

@String(ECCV  = {ECCV})

@article{ravi2024sam2,
  title={SAM 2: Segment Anything in Images and Videos},
  author={Ravi, Nikhila and Gabeur, Valentin and Hu, Yuan-Ting and Hu, Ronghang and Ryali, Chaitanya and Ma, Tengyu and Khedr, Haitham and R{\"a}dle, Roman and Rolland, Chloe and Gustafson, Laura and Mintun, Eric and Pan, Junting and Alwala, Kalyan Vasudev and Carion, Nicolas and Wu, Chao-Yuan and Girshick, Ross and Doll{\'a}r, Piotr and Feichtenhofer, Christoph},
  journal={arXiv preprint arXiv:2408.00714},
  url={https://arxiv.org/abs/2408.00714},
  year={2024}
}

@inproceedings{yeshwanth2023scannet++,
  title={Scannet++: A high-fidelity dataset of 3d indoor scenes},
  author={Yeshwanth, Chandan and Liu, Yueh-Cheng and Nie{\ss}ner, Matthias and Dai, Angela},
  booktitle={Proceedings of the IEEE/CVF International Conference on Computer Vision},
  pages={12--22},
  year={2023}
}

@inproceedings{wang2025vggt,
  title={Vggt: Visual geometry grounded transformer},
  author={Wang, Jianyuan and Chen, Minghao and Karaev, Nikita and Vedaldi, Andrea and Rupprecht, Christian and Novotny, David},
  booktitle={Proceedings of the Computer Vision and Pattern Recognition Conference},
  pages={5294--5306},
  year={2025}
}

@inproceedings{schoenberger2016sfm,
    author={Sch\"{o}nberger, Johannes Lutz and Frahm, Jan-Michael},
    title={Structure-from-Motion Revisited},
    booktitle={Conference on Computer Vision and Pattern Recognition (CVPR)},
    year={2016},
}

@inproceedings{wang2024dust3r,
  title={Dust3r: Geometric 3d vision made easy},
  author={Wang, Shuzhe and Leroy, Vincent and Cabon, Yohann and Chidlovskii, Boris and Revaud, Jerome},
  booktitle={Proceedings of the IEEE/CVF Conference on Computer Vision and Pattern Recognition},
  pages={20697--20709},
  year={2024}
}

@inproceedings{dai2017scannet,
  title={Scannet: Richly-annotated 3d reconstructions of indoor scenes},
  author={Dai, Angela and Chang, Angel X and Savva, Manolis and Halber, Maciej and Funkhouser, Thomas and Nie{\ss}ner, Matthias},
  booktitle={Proceedings of the IEEE conference on computer vision and pattern recognition},
  pages={5828--5839},
  year={2017}
}

@article{oquab2023dinov2,
  title={Dinov2: Learning robust visual features without supervision},
  author={Oquab, Maxime and Darcet, Timoth{\'e}e and Moutakanni, Th{\'e}o and Vo, Huy and Szafraniec, Marc and Khalidov, Vasil and Fernandez, Pierre and Haziza, Daniel and Massa, Francisco and El-Nouby, Alaaeldin and others},
  journal={arXiv preprint arXiv:2304.07193},
  year={2023}
}

@inproceedings{radford2021clip,
  title={Learning transferable visual models from natural language supervision},
  author={Radford, Alec and Kim, Jong Wook and Hallacy, Chris and Ramesh, Aditya and Goh, Gabriel and Agarwal, Sandhini and Sastry, Girish and Askell, Amanda and Mishkin, Pamela and Clark, Jack and others},
  booktitle={International conference on machine learning},
  pages={8748--8763},
  year={2021},
  organization={PmLR}
}

@misc{wang2025pi3,
      title={$\pi^3$: Permutation-Equivariant Visual Geometry Learning}, 
      author={Yifan Wang and Jianjun Zhou and Haoyi Zhu and Wenzheng Chang and Yang Zhou and Zizun Li and Junyi Chen and Jiangmiao Pang and Chunhua Shen and Tong He},
      year={2025},
      eprint={2507.13347},
      archivePrefix={arXiv},
      primaryClass={cs.CV},
      url={https://arxiv.org/abs/2507.13347}, 
}

@inproceedings{yang2025fast3r,
  title={Fast3r: Towards 3d reconstruction of 1000+ images in one forward pass},
  author={Yang, Jianing and Sax, Alexander and Liang, Kevin J and Henaff, Mikael and Tang, Hao and Cao, Ang and Chai, Joyce and Meier, Franziska and Feiszli, Matt},
  booktitle={Proceedings of the Computer Vision and Pattern Recognition Conference},
  pages={21924--21935},
  year={2025}
}

@article{pont20172017,
  title={The 2017 davis challenge on video object segmentation},
  author={Pont-Tuset, Jordi and Perazzi, Federico and Caelles, Sergi and Arbel{\'a}ez, Pablo and Sorkine-Hornung, Alex and Van Gool, Luc},
  journal={arXiv preprint arXiv:1704.00675},
  year={2017}
}

@article{cheng2021rethinking,
  title={Rethinking space-time networks with improved memory coverage for efficient video object segmentation},
  author={Cheng, Ho Kei and Tai, Yu-Wing and Tang, Chi-Keung},
  journal={Advances in Neural Information Processing Systems},
  volume={34},
  pages={11781--11794},
  year={2021}
}

@inproceedings{MOSE,
  title={{MOSE}: A New Dataset for Video Object Segmentation in Complex Scenes},
  author={Ding, Henghui and Liu, Chang and He, Shuting and Jiang, Xudong and Torr, Philip HS and Bai, Song},
  booktitle={ICCV},
  year={2023}
}

@inproceedings{bekuzarov2023xmem++,
  title={Xmem++: Production-level video segmentation from few annotated frames},
  author={Bekuzarov, Maksym and Bermudez, Ariana and Lee, Joon-Young and Li, Hao},
  booktitle={Proceedings of the IEEE/CVF International Conference on Computer Vision},
  pages={635--644},
  year={2023}
}

@inproceedings{cheng2024putting,
  title={Putting the object back into video object segmentation},
  author={Cheng, Ho Kei and Oh, Seoung Wug and Price, Brian and Lee, Joon-Young and Schwing, Alexander},
  booktitle={Proceedings of the IEEE/CVF Conference on Computer Vision and Pattern Recognition},
  pages={3151--3161},
  year={2024}
}

@article{zhang2023joint,
  title={Joint modeling of feature, correspondence, and a compressed memory for video object segmentation},
  author={Zhang, Jiaming and Cui, Yutao and Wu, Gangshan and Wang, Limin},
  journal={arXiv preprint arXiv:2308.13505},
  year={2023}
}

@inproceedings{li2022recurrent,
  title={Recurrent dynamic embedding for video object segmentation},
  author={Li, Mingxing and Hu, Li and Xiong, Zhiwei and Zhang, Bang and Pan, Pan and Liu, Dong},
  booktitle={Proceedings of the IEEE/CVF Conference on Computer Vision and Pattern Recognition},
  pages={1332--1341},
  year={2022}
}

@inproceedings{wu2023scalable,
  title={Scalable video object segmentation with simplified framework},
  author={Wu, Qiangqiang and Yang, Tianyu and Wu, Wei and Chan, Antoni B},
  booktitle={Proceedings of the IEEE/CVF International Conference on Computer Vision},
  pages={13879--13889},
  year={2023}
}

@inproceedings{wang2023look,
  title={Look before you match: Instance understanding matters in video object segmentation},
  author={Wang, Junke and Chen, Dongdong and Wu, Zuxuan and Luo, Chong and Tang, Chuanxin and Dai, Xiyang and Zhao, Yucheng and Xie, Yujia and Yuan, Lu and Jiang, Yu-Gang},
  booktitle={Proceedings of the IEEE/CVF conference on computer vision and pattern recognition},
  pages={2268--2278},
  year={2023}
}

@article{li2023semantic,
  title={Semantic-sam: Segment and recognize anything at any granularity},
  author={Li, Feng and Zhang, Hao and Sun, Peize and Zou, Xueyan and Liu, Shilong and Yang, Jianwei and Li, Chunyuan and Zhang, Lei and Gao, Jianfeng},
  journal={arXiv preprint arXiv:2307.04767},
  year={2023}
}

@article{ma2024segment,
  title={Segment anything in medical images},
  author={Ma, Jun and He, Yuting and Li, Feifei and Han, Lin and You, Chenyu and Wang, Bo},
  journal={Nature Communications},
  volume={15},
  number={1},
  pages={654},
  year={2024},
  publisher={Nature Publishing Group UK London}
}

@inproceedings{kirillov2023segment,
  title={Segment anything},
  author={Kirillov, Alexander and Mintun, Eric and Ravi, Nikhila and Mao, Hanzi and Rolland, Chloe and Gustafson, Laura and Xiao, Tete and Whitehead, Spencer and Berg, Alexander C and Lo, Wan-Yen and others},
  booktitle={Proceedings of the IEEE/CVF International Conference on Computer Vision},
  pages={4015--4026},
  year={2023}
}

@inproceedings{cheng2022xmem,
  title={Xmem: Long-term video object segmentation with an atkinson-shiffrin memory model},
  author={Cheng, Ho Kei and Schwing, Alexander G},
  booktitle={European Conference on Computer Vision},
  pages={640--658},
  year={2022},
  organization={Springer}
}

@article{yang2021aot,
  title={Associating objects with transformers for video object segmentation},
  author={Yang, Zongxin and Wei, Yunchao and Yang, Yi},
  journal={Advances in Neural Information Processing Systems},
  volume={34},
  pages={2491--2502},
  year={2021}
}

@article{yang2022deaot,
  title={Decoupling features in hierarchical propagation for video object segmentation},
  author={Yang, Zongxin and Yang, Yi},
  journal={Advances in Neural Information Processing Systems},
  volume={35},
  pages={36324--36336},
  year={2022}
}

@inproceedings{wu2022lreferformer,
  title={Language as queries for referring video object segmentation},
  author={Wu, Jiannan and Jiang, Yi and Sun, Peize and Yuan, Zehuan and Luo, Ping},
  booktitle={Proceedings of the IEEE/CVF Conference on Computer Vision and Pattern Recognition},
  pages={4974--4984},
  year={2022}
}

@inproceedings{han2023html,
  title={Html: Hybrid temporal-scale multimodal learning framework for referring video object segmentation},
  author={Han, Mingfei and Wang, Yali and Li, Zhihui and Yao, Lina and Chang, Xiaojun and Qiao, Yu},
  booktitle={Proceedings of the IEEE/CVF International Conference on Computer Vision},
  pages={13414--13423},
  year={2023}
}

@inproceedings{miao2023sgmg,
  title={Spectrum-guided multi-granularity referring video object segmentation},
  author={Miao, Bo and Bennamoun, Mohammed and Gao, Yongsheng and Mian, Ajmal},
  booktitle={Proceedings of the IEEE/CVF International Conference on Computer Vision},
  pages={920--930},
  year={2023}
}

@article{liang2025referdino,
  title={Referdino: Referring video object segmentation with visual grounding foundations},
  author={Liang, Tianming and Lin, Kun-Yu and Tan, Chaolei and Zhang, Jianguo and Zheng, Wei-Shi and Hu, Jian-Fang},
  journal={arXiv preprint arXiv:2501.14607},
  year={2025}
}

@inproceedings{cheng2023deva,
  title={Tracking anything with decoupled video segmentation},
  author={Cheng, Ho Kei and Oh, Seoung Wug and Price, Brian and Schwing, Alexander and Lee, Joon-Young},
  booktitle={Proceedings of the IEEE/CVF International Conference on Computer Vision},
  pages={1316--1326},
  year={2023}
}

@inproceedings{ScanRefer,
  title={Scanrefer: 3d object localization in rgb-d scans using natural language},
  author={Chen, Dave Zhenyu and Chang, Angel X and Nie{\ss}ner, Matthias},
  booktitle=ECCV,
  year={2020},
}

@inproceedings{ReferIt3D,
  title={Referit3d: Neural listeners for fine-grained 3d object identification in real-world scenes},
  author={Achlioptas, Panos and Abdelreheem, Ahmed and Xia, Fei and Elhoseiny, Mohamed and Guibas, Leonidas},
  booktitle=ECCV,
  year={2020},
}

@inproceedings{butd-detr,
  title={Bottom up top down detection transformers for language grounding in images and point clouds},
  author={Jain, Ayush and Gkanatsios, Nikolaos and Mediratta, Ishita and Fragkiadaki, Katerina},
  booktitle=ECCV,
  year={2022},
}

@inproceedings{instancerefer,
  title={Instancerefer: Cooperative holistic understanding for visual grounding on point clouds through instance multi-level contextual referring},
  author={Yuan, Zhihao and Yan, Xu and Liao, Yinghong and Zhang, Ruimao and Wang, Sheng and Li, Zhen and Cui, Shuguang},
  booktitle=ICCV,
  year={2021}
}

@inproceedings{3d-vista,
  title={3d-vista: Pre-trained transformer for 3d vision and text alignment},
  author={Zhu, Ziyu and Ma, Xiaojian and Chen, Yixin and Deng, Zhidong and Huang, Siyuan and Li, Qing},
  booktitle=ICCV,
  year={2023}
}

@inproceedings{LLM-Grounder,
  title={Llm-grounder: Open-vocabulary 3d visual grounding with large language model as an agent},
  author={Yang, Jianing and Chen, Xuweiyi and Qian, Shengyi and Madaan, Nikhil and Iyengar, Madhavan and Fouhey, David F and Chai, Joyce},
  booktitle={ICRA},
  year={2024}
}

@inproceedings{Languagerefer,
  title={Languagerefer: Spatial-language model for 3d visual grounding},
  author={Roh, Junha and Desingh, Karthik and Farhadi, Ali and Fox, Dieter},
  booktitle={CoRL},
  year={2022},
}

@inproceedings{OpenScene,
  title={Openscene: 3d scene understanding with open vocabularies},
  author={Peng, Songyou and Genova, Kyle and Jiang, Chiyu and Tagliasacchi, Andrea and Pollefeys, Marc and Funkhouser, Thomas and others},
  booktitle=CVPR,
}

@inproceedings{wang2025continuous,
  title={Continuous 3d perception model with persistent state},
  author={Wang, Qianqian and Zhang, Yifei and Holynski, Aleksander and Efros, Alexei A and Kanazawa, Angjoo},
  booktitle={Proceedings of the Computer Vision and Pattern Recognition Conference},
  pages={10510--10522},
  year={2025}
}

@inproceedings{zhang2025flare,
  title={Flare: Feed-forward geometry, appearance and camera estimation from uncalibrated sparse views},
  author={Zhang, Shangzhan and Wang, Jianyuan and Xu, Yinghao and Xue, Nan and Rupprecht, Christian and Zhou, Xiaowei and Shen, Yujun and Wetzstein, Gordon},
  booktitle={Proceedings of the Computer Vision and Pattern Recognition Conference},
  pages={21936--21947},
  year={2025}
}

@book{hartley2003multiple,
  title={Multiple view geometry in computer vision},
  author={Hartley, Richard and Zisserman, Andrew},
  year={2003},
  publisher={Cambridge university press}
}

@inproceedings{pan2024global,
  title={Global structure-from-motion revisited},
  author={Pan, Linfei and Bar{\'a}th, D{\'a}niel and Pollefeys, Marc and Sch{\"o}nberger, Johannes L},
  booktitle={European Conference on Computer Vision},
  pages={58--77},
  year={2024},
  organization={Springer}
}

@article{furukawa2015multi,
  title={Multi-view stereo: A tutorial},
  author={Furukawa, Yasutaka and Hern{\'a}ndez, Carlos and others},
  journal={Foundations and trends{\textregistered} in Computer Graphics and Vision},
  volume={9},
  number={1-2},
  pages={1--148},
  year={2015},
  publisher={Now Publishers, Inc.}
}

@inproceedings{schonberger2016pixelwise,
  title={Pixelwise view selection for unstructured multi-view stereo},
  author={Sch{\"o}nberger, Johannes L and Zheng, Enliang and Frahm, Jan-Michael and Pollefeys, Marc},
  booktitle={European conference on computer vision},
  pages={501--518},
  year={2016},
  organization={Springer}
}

@article{yuan2025sa2va,
  title={Sa2va: Marrying sam2 with llava for dense grounded understanding of images and videos},
  author={Yuan, Haobo and Li, Xiangtai and Zhang, Tao and Huang, Zilong and Xu, Shilin and Ji, Shunping and Tong, Yunhai and Qi, Lu and Feng, Jiashi and Yang, Ming-Hsuan},
  journal={arXiv preprint arXiv:2501.04001},
  year={2025}
}

@inproceedings{seo2020ref-youtubevos,
  title={Urvos: Unified referring video object segmentation network with a large-scale benchmark},
  author={Seo, Seonguk and Lee, Joon-Young and Han, Bohyung},
  booktitle={European conference on computer vision},
  pages={208--223},
  year={2020},
  organization={Springer}
}

@article{baruch2021arkitscenes,
  title={Arkitscenes: A diverse real-world dataset for 3d indoor scene understanding using mobile rgb-d data},
  author={Baruch, Gilad and Chen, Zhuoyuan and Dehghan, Afshin and Dimry, Tal and Feigin, Yuri and Fu, Peter and Gebauer, Thomas and Joffe, Brandon and Kurz, Daniel and Schwartz, Arik and others},
  journal={arXiv preprint arXiv:2111.08897},
  year={2021}
}

@article{zhou2025omniworld,
  title={OmniWorld: A Multi-Domain and Multi-Modal Dataset for 4D World Modeling},
  author={Zhou, Yang and Wang, Yifan and Zhou, Jianjun and Chang, Wenzheng and Guo, Haoyu and Li, Zizun and Ma, Kaijing and Li, Xinyue and Wang, Yating and Zhu, Haoyi and others},
  journal={arXiv preprint arXiv:2509.12201},
  year={2025}
}

@article{arnaud2025locate3d,
  title={Locate 3D: Real-World Object Localization via Self-Supervised Learning in 3D},
  author={Arnaud, Sergio and McVay, Paul and Martin, Ada and Majumdar, Arjun and Jatavallabhula, Krishna Murthy and Thomas, Phillip and Partsey, Ruslan and Dugas, Daniel and Gejji, Abha and Sax, Alexander and others},
  journal={arXiv preprint arXiv:2504.14151},
  year={2025}
}

@inproceedings{leroy2024grounding,
  title={Grounding image matching in 3d with mast3r},
  author={Leroy, Vincent and Cabon, Yohann and Revaud, J{\'e}r{\^o}me},
  booktitle={European Conference on Computer Vision},
  pages={71--91},
  year={2024},
  organization={Springer}
}

@article{zhang2024monst3r,
  title={Monst3r: A simple approach for estimating geometry in the presence of motion},
  author={Zhang, Junyi and Herrmann, Charles and Hur, Junhwa and Jampani, Varun and Darrell, Trevor and Cole, Forrester and Sun, Deqing and Yang, Ming-Hsuan},
  journal={arXiv preprint arXiv:2410.03825},
  year={2024}
}

@article{bozic2021transformerfusion,
  title={Transformerfusion: Monocular rgb scene reconstruction using transformers},
  author={Bozic, Aljaz and Palafox, Pablo and Thies, Justus and Dai, Angela and Nie{\ss}ner, Matthias},
  journal={Advances in Neural Information Processing Systems},
  volume={34},
  pages={1403--1414},
  year={2021}
}

@article{geiger2013kitti,
  title={Vision meets robotics: The kitti dataset},
  author={Geiger, Andreas and Lenz, Philip and Stiller, Christoph and Urtasun, Raquel},
  journal={The international journal of robotics research},
  volume={32},
  number={11},
  pages={1231--1237},
  year={2013},
  publisher={Sage Publications Sage UK: London, England}
}

@inproceedings{silberman2012indoor,
  title={Indoor segmentation and support inference from rgbd images},
  author={Silberman, Nathan and Hoiem, Derek and Kohli, Pushmeet and Fergus, Rob},
  booktitle={European conference on computer vision},
  pages={746--760},
  year={2012},
  organization={Springer}
}

@inproceedings{botach2022end,
  title={End-to-end referring video object segmentation with multimodal transformers},
  author={Botach, Adam and Zheltonozhskii, Evgenii and Baskin, Chaim},
  booktitle={Proceedings of the IEEE/CVF Conference on Computer Vision and Pattern Recognition},
  pages={4985--4995},
  year={2022}
}

@article{huang2025surprise3d,
  title={SURPRISE3D: A Dataset for Spatial Understanding and Reasoning in Complex 3D Scenes},
  author={Huang, Jiaxin and Li, Ziwen and Zhang, Hanlve and Chen, Runnan and He, Xiao and Guo, Yandong and Wang, Wenping and Liu, Tongliang and Gong, Mingming},
  journal={arXiv preprint arXiv:2507.07781},
  year={2025}
}

@article{zhang2025SPAR,
  title={From flatland to space: Teaching vision-language models to perceive and reason in 3d},
  author={Zhang, Jiahui and Chen, Yurui and Zhou, Yanpeng and Xu, Yueming and Huang, Ze and Mei, Jilin and Chen, Junhui and Yuan, Yu-Jie and Cai, Xinyue and Huang, Guowei and others},
  journal={arXiv preprint arXiv:2503.22976},
  year={2025}
}

@inproceedings{dpt,
  title={Vision transformers for dense prediction},
  author={Ranftl, Ren{\'e} and Bochkovskiy, Alexey and Koltun, Vladlen},
  booktitle={Proceedings of the IEEE/CVF international conference on computer vision},
  pages={12179--12188},
  year={2021}
}

@inproceedings{LMK,
  title={Spatial cognition from egocentric video: Out of sight, not out of mind},
  author={Plizzari, Chiara and Goel, Shubham and Perrett, Toby and Chalk, Jacob and Kanazawa, Angjoo and Damen, Dima},
  booktitle={2025 International Conference on 3D Vision (3DV)},
  pages={1211--1221},
  year={2025},
  organization={IEEE}
}

@inproceedings{3dtracking,
  title={3d-aware instance segmentation and tracking in egocentric videos},
  author={Bhalgat, Yash and Tschernezki, Vadim and Laina, Iro and Henriques, Jo{\~a}o F and Vedaldi, Andrea and Zisserman, Andrew},
  booktitle={Proceedings of the Asian Conference on Computer Vision},
  pages={2562--2578},
  year={2024}
}

@inproceedings{ding2023mevis,
  title={Mevis: A large-scale benchmark for video segmentation with motion expressions},
  author={Ding, Henghui and Liu, Chang and He, Shuting and Jiang, Xudong and Loy, Chen Change},
  booktitle={Proceedings of the IEEE/CVF international conference on computer vision},
  pages={2694--2703},
  year={2023}
}

@article{vlm-grounder,
  title={Vlm-grounder: A vlm agent for zero-shot 3d visual grounding},
  author={Xu, Runsen and Huang, Zhiwei and Wang, Tai and Chen, Yilun and Pang, Jiangmiao and Lin, Dahua},
  journal={arXiv preprint arXiv:2410.13860},
  year={2024}
}
\bibliographystyle{icml2026}

\newpage
\appendix
\onecolumn

\section{InsTrack Dataset Generation}
\label{appendix:pipeline}

We create an automatic annotation pipeline to generate multi-modal prompts and instance-consistent masks for each frame from the ScanNet++~\cite{yeshwanth2023scannet++} dataset. The pipeline consists of three stages:

\noindent\textit{(1) Image Subsampling.} Since the video frames provided by ScanNet++ contain a significant amount of redundant information, we adopt the image filtering method from SPAR~\cite{zhang2025SPAR}, which leverages camera poses to reduce redundant images with high similarity. This approach effectively filters out approximately 80\% of the redundant images, ensuring that only images with sufficiently distinct poses are retained, thereby achieving the best possible scene coverage with a limited number of frames.

\vspace{0.5mm}

\noindent\textit{(2) 2D Instance Mask Generation.} Since ScanNet++ only officially provides 3D instance mesh annotations, we first rasterize the 3D meshes onto 2D images and save the 2D-3D mappings (pixel-to-face), which map image pixels to face indices. Then, we utilize this rasterization to obtain the instance annotations on the 2D images, thus constructing the corresponding instance ID map.

\vspace{0.5mm}

\noindent\textit{(3) Multi-modal Prompting.} We curate a diverse set of visual and text prompts to guide instance tracking. For visual prompts, we select up to 20 objects per scene that appear in at least five subsampled frames. From the images containing a target object, we randomly choose one as the reference frame, where the instance mask provides the \textit{mask prompt}.  
To enhance prompt diversity, we further sample five points within the mask as \textit{point prompts}, and use the tight bounding box of the mask as the \textit{box prompt}. For text prompts, we integrate a subset of annotations from L3DD~\cite{arnaud2025locate3d} and SURPRISE3D~\cite{huang2025surprise3d}, encompassing human-annotated referring expressions and complex language-guided queries that involve spatial, commonsense, and intentional reasoning.

\section{Training Details}
\label{appendix:training_details}

To ensure robustness and broad applicability across indoor/outdoor and static/dynamic scenarios, we curate a large-scale three-part corpus to train \modelname.
The corpus comprises (i) video object segmentation datasets: VOS (DAVIS, MOSE, YouTubeVOS), SA-V~\cite{ravi2024sam2}, Ref-SAV~\cite{yuan2025sa2va}, and Ref-YTVOS~\cite{seo2020ref-youtubevos}, (ii) reconstruction datasets: ScanNet++~\cite{yeshwanth2023scannet++}, ScanNet~\cite{dai2017scannet}, ARKitScenes~\cite{baruch2021arkitscenes}, and OmniWorld~\cite{zhou2025omniworld}, and (iii) our joint segmentation–reconstruction dataset: InsTrack.
We initialize the model with weights from Pi3~\cite{wang2025pi3} and fine-tune on the aggregated corpus. Training is conducted on 64 NVIDIA A100 GPUs using AdamW.
We adopt differential learning rates for different model components: $6\times 10^{-6}$ for the cross-modal spatial encoder and $1\times 10^{-5}$ for the visual prompt encoder, geometry decoder, and mask decoder.
We use the CLIP encoder~\cite{radford2021clip} as the text prompt encoder and freeze it throughout training.
To stabilize optimization, the data sampler ensures that each mini-batch contains only one data type.
Our objective combines segmentation and geometry losses with weights $\lambda_{\text{seg}}=2.0$ and $\lambda_{\text{geo}}=1.0$, and the component weights within $\mathcal{L}_{\text{geo}}$ keeps the same with Pi3. The learning objective math formulation is provided in Appendix~\ref{app:losses}.

\section{Architecture Details}

\paragraph{Frame-wise ViT Encoder.} As outlined in the main paper, each input view is embedded into a sequence of patch-level vision tokens using DINOv2~\cite{oquab2023dinov2}. To adapt the encoder for dense prediction tasks, we adopt a feature reassembly strategy inspired by DPT~\cite{dpt}. Specifically, we extract features from transformer layers $l = \{5, 12\}$. Given that the ViT architecture is isotropic (i.e., features across layers share the same resolution), we employ convolutional layers to project the channel dimensions from 1024 to 32 and 64, respectively. Subsequently, these features are upsampled by factors of $4\times$ and $2\times$ to restore spatial resolution. Finally, these multi-scale features are integrated into the mask decoder via skip connections.

\paragraph{Task Modes in Cross-Modal Spatial Encoder.}

\begin{figure}[!ht]
  \centering
  \includegraphics[width=0.8\linewidth]{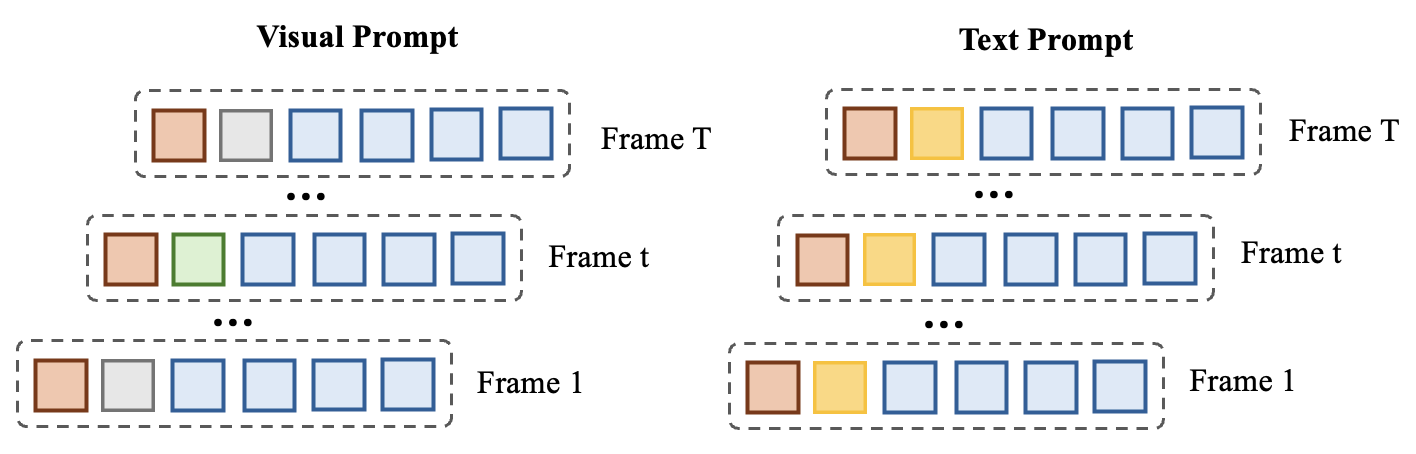}
  \vspace{-2ex}
  \caption{\textbf{Modes in Cross-Model Spatial Encoder.} }
  \vspace{-2ex}
  \label{fig:mode}
\end{figure}

We utilize Fig.~\ref{fig:mode} to illustrate the modes of our cross-modal spatial encoder when handling different types of prompts.
\textit{Visual Prompts (Left)}: As depicted in the left panel, when a visual prompt is provided on frame $t$, it is projected into a \textcolor{cvprgreen}{visual prompt embedding} via the visual prompt encoder. This embedding is then inserted between the register tokens and the vision tokens extracted by DINOv2. For other frames without visual prompts, a \textcolor{cvprgray}{learnable placeholder embedding} is inserted at the corresponding position to maintain structural consistency. \textit{Text Prompts (Right)}: In the case of text prompts, as shown in the right panel, the \textcolor{cvprorange}{text prompt embeddings} are inserted between the register tokens and vision tokens across all frames.

\begin{figure}[!ht]
  \centering
  \includegraphics[width=0.8\linewidth]{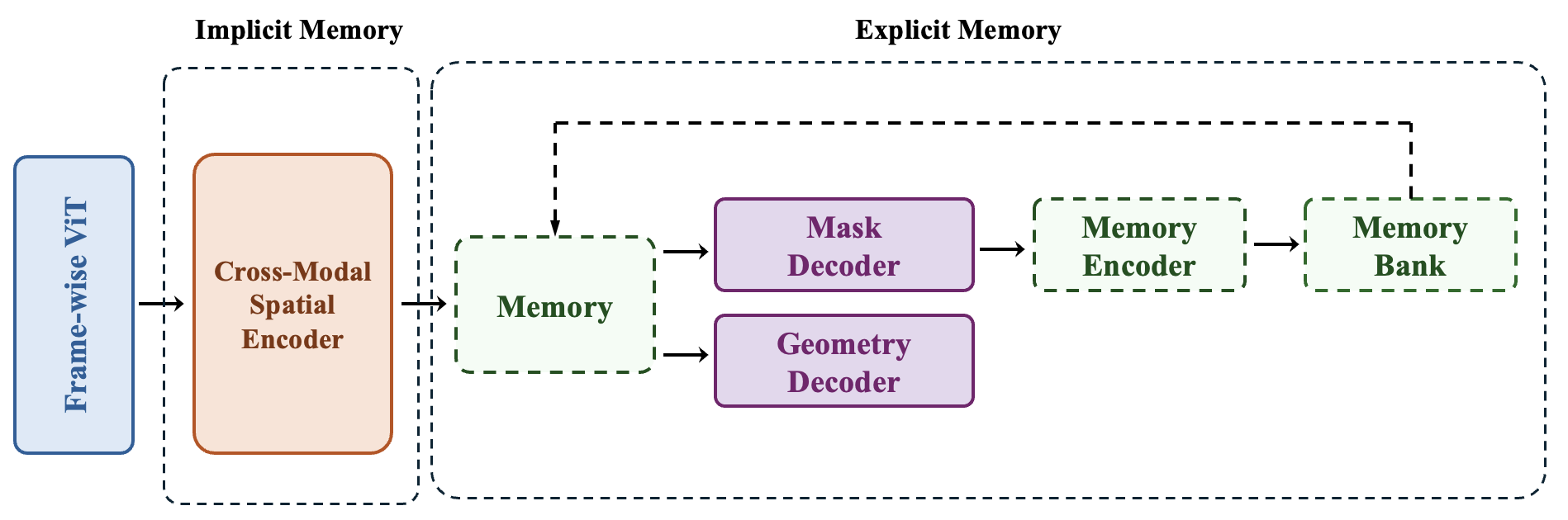}
  \vspace{-2ex}
  \caption{\textbf{Implicit and Explicit Memory Combination Architecture.} }
  \vspace{-2ex}
  \label{fig:explicit_model}
\end{figure}

\paragraph{Architecture of Implicit and Explicit Combination.}
\label{appendix:combine}

In this section, we present the combination of implicit and explicit memory model architecture within our ablation study. Building upon the original \modelname architecture, we incorporate the memory mechanism from SAM2~\cite{ravi2024sam2} by introducing additional memory encoder and memory bank modules (highlighted by the green dashed box in Fig.~\ref{fig:explicit_model}). Specifically, the memory encoder generates memory features by downsampling the output mask via a convolutional module. This representation is summed element-wise with the unconditioned frame embedding from the image encoder, followed by lightweight convolutional layers to fuse the information. The memory bank preserves temporal context by maintaining a First-In-First-Out (FIFO) queue of memories from the $N$ most recent frames. Consequently, within this explicit memory architecture, each frame prediction is conditioned not only on the current frame features but also on historical mask prediction results.

\begin{figure*}[!ht]
  \centering
  \includegraphics[width=1\linewidth]{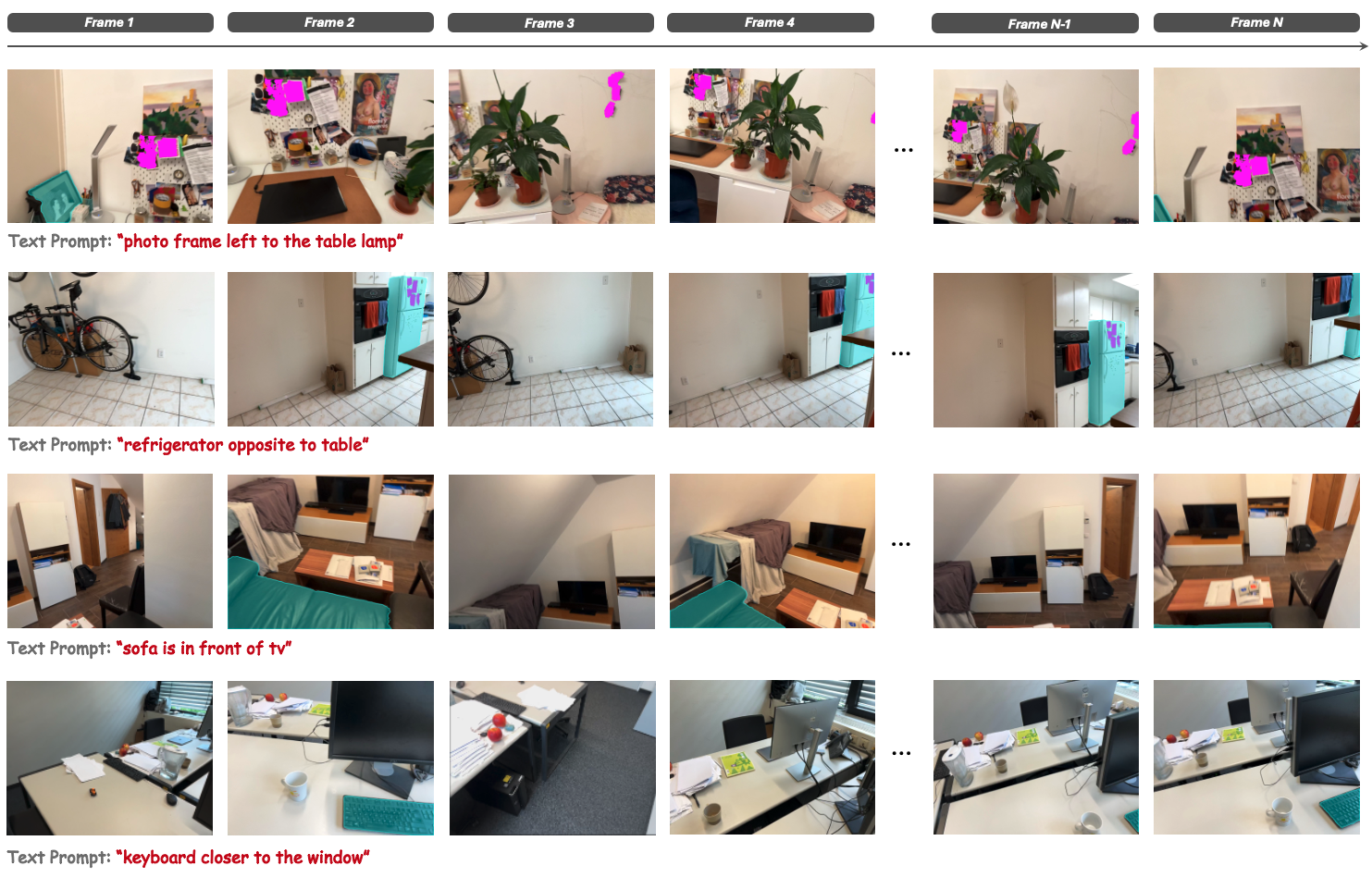}
  \caption{\textbf{Qualitative results on InsTrack validation set.} These samples typically focus on the sub-region of the large scene, and the text prompts typically consist of simple and short spatial relationships. }
  \label{fig:viz_1}
\end{figure*}

\section{More Visualization Results}


We first present more visualization results of \modelname on the InsTrack validation set, which are primarily derived from L3DD~\cite{arnaud2025locate3d}. It is worth noting that the input images for these samples do not necessarily cover the entire scene; rather, they may focus on selected sub-regions within a larger environment. In this setting, our method demonstrates preliminary capabilities in spatial reasoning. For example, in the scene shown in the bottom row of Fig.~\ref{fig:viz_1}, despite the presence of two keyboards, the model successfully identifies \textit{``the keyboard closer to the window''} from the one that appeared in the first frame. Furthermore, the model exhibits remarkable cross-view consistency even across significant viewpoint changes. 

\begin{figure*}[!t]
  \centering
  \includegraphics[width=0.9\linewidth]{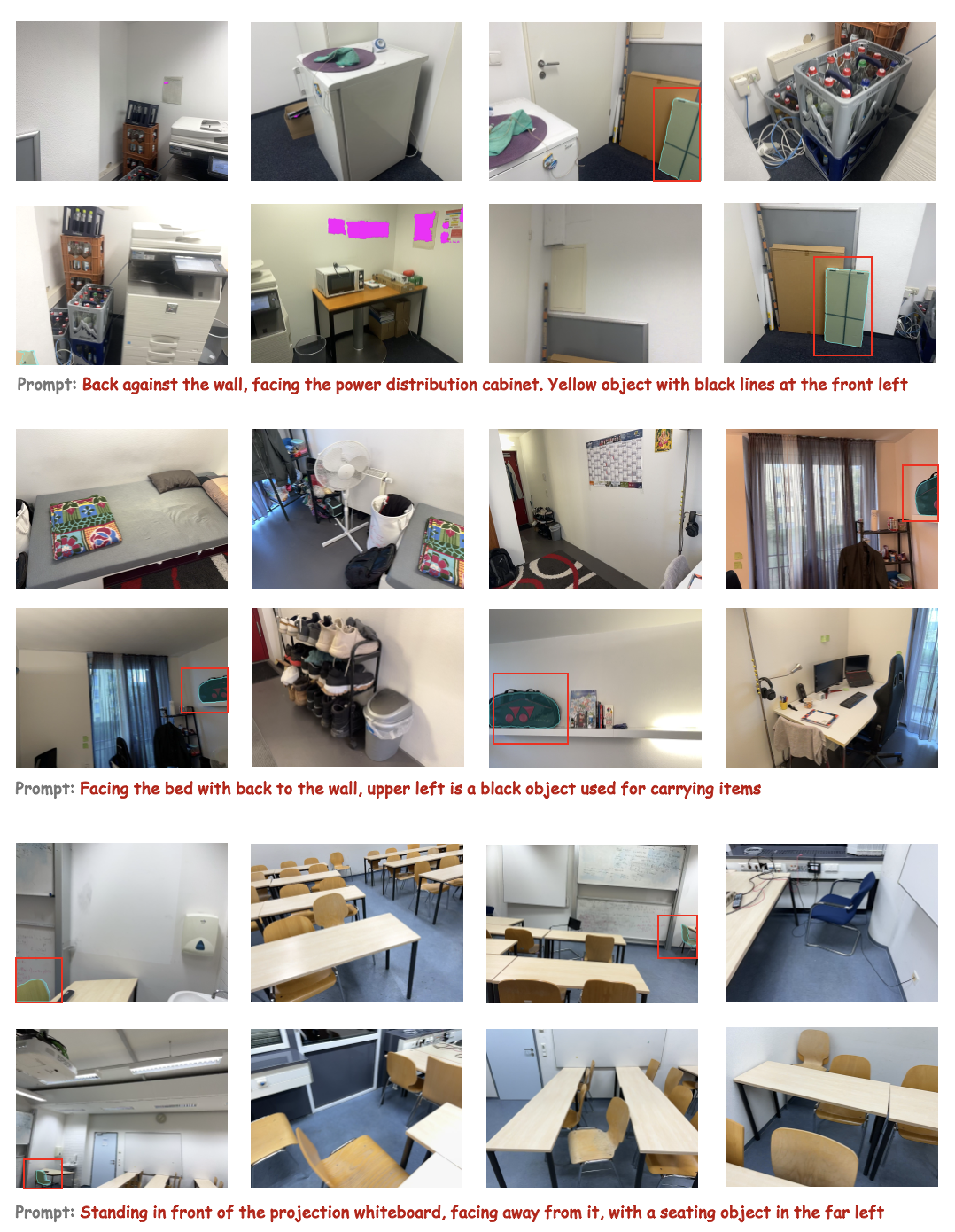}
  \caption{\textbf{Qualitative results on InsTrack validation set.} These samples cover the entire large scenes and the text prompts are much more complicated.}
  \label{fig:viz_2}
\end{figure*}

We further extend our analysis to more challenging scenarios, as visualized in Fig.~\ref{fig:viz_2}. In contrast to the previous samples, these images typically capture comprehensive views of the entire scene and involve significantly more complicated text prompts. Beyond simple pairwise spatial relationships, these prompts require \textbf{commonsense knowledge} (e.g., \textit{object used for carrying items}), \textbf{logical reasoning}, and \textbf{spatial imagination} (e.g., \textit{facing the bed with back to the wall}, \textit{standing in front of the projection whiteboard}). Despite this complexity, \modelname accurately reasons through the instructions and consistently localizes the target objects across different frames. A notable example is shown in the bottom row: although the scene contains numerous chairs that satisfy the general description of ``\textit{a seating object}'', our model successfully disambiguates and identifies the specific chair that strictly satisfies all conditions. The segmented target objects are highlighted with red bounding boxes in the image.

\begin{figure*}[!t]
  \centering
  \includegraphics[width=1\linewidth]{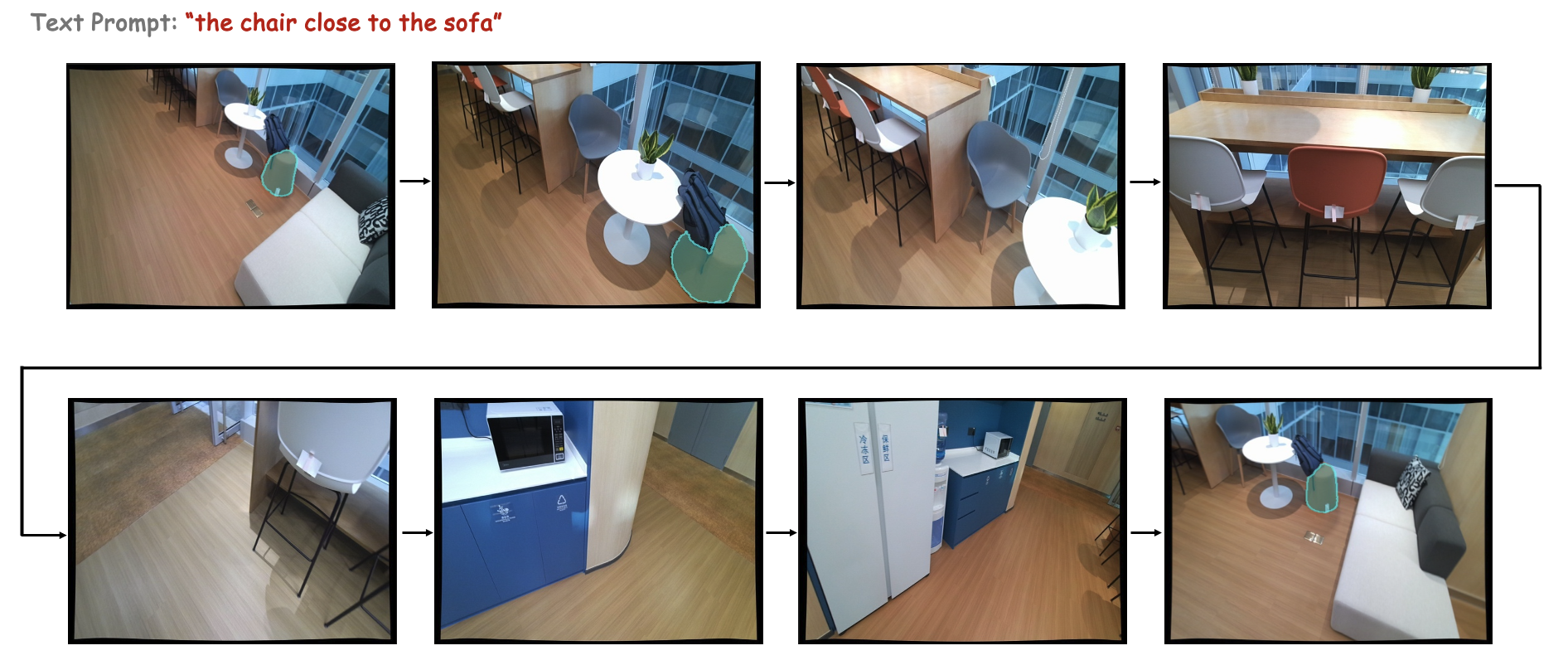}
  \caption{\textbf{Illustration of Explore-and-Revisit Trajectory.} Our \modelname maintains consistent object identity over long-term exploration, successfully re-recognizing targets during revisits.}
  \label{fig:mem}
\end{figure*}

\section{More Spatial Memory Demonstration}

\paragraph{Explore and Revisit.} To further demonstrate the spatial memory capabilities of our model, we curated a set of long-term scenarios featuring an explore-and-revisit trajectory. In these sequences, the camera starts at a specific location, explores the environment for an extended duration, and subsequently returns to the initial scene. This setup simulates a realistic exploration mechanism. As shown in the Fig.~\ref{fig:mem},  our model accurately comprehends the textual spatial descriptions in the initial frames (Frame 1 and Frame 2). Crucially, after a prolonged period of absence and exploration, when the camera returns to the starting point, our model successfully recalls the scene, accurately recognizing and segmenting the target objects despite the temporal gap.

\begin{table}[t]
\centering
\resizebox{0.6\columnwidth}{!}{
\begin{tabular}{lcc}
\toprule
\textbf{Benchmark}  & \textbf{Original Temporal Order} & \textbf{Shuffled Temporal Order} \\
\midrule
InsTrack \textit{Visual} & \textbf{75.8} & 75.4 \\
InsTrack \textit{Text}   & \textbf{72.3 }& 70.1 \\ 
\bottomrule
\end{tabular}
}
\vspace{4pt}
\caption{Ablation on robustness to temporal permutation for InsTrack \textit{Visual} and \textit{Text}.}
\label{tab:robustness_instrack}
\end{table}

\paragraph{Robustness to Temporal Permutation.} Although our InsTrack dataset ensures significant viewpoint changes between frames, the sequences remain temporally consistent. To rigorously verify our model's spatial memory capabilities, we conducted an experiment on the InsTrack validation set where we randomly shuffled the input image sequences. This operation removes cues derived from visual temporal continuity, forcing the model to identify object consistency without relying on smooth temporal transitions.

Results in Tab.~\ref{tab:robustness_instrack} indicate that performance remains largely consistent with visual prompts. However, we observe a slight performance degradation with text prompts; we hypothesize that this is because certain text descriptions may be implicitly correlated with viewpoint continuity, making them sensitive to temporal disorder. Despite this, the model demonstrates overall robustness to temporal reordering, effectively exhibiting permutation equivariance (see Tab.~\ref{tab:robustness_instrack}). These findings confirm that our model relies primarily on spatial consistency across different viewpoints rather than temporal proximity in static scenes, validating the effectiveness of our spatial memory mechanism.

\section{Learning Objective Details}
\label{app:losses}

To provide a rigorous mathematical foundation for the training process, we detail the formulation of each loss component. The total loss $\mathcal{L}$ is a weighted combination of the segmentation objective and the geometry objective:
\begin{equation}
    \mathcal{L} = \lambda_{\text{seg}} \mathcal{L}_{\text{seg}} + \lambda_{\text{geo}} \mathcal{L}_{\text{geo}}
\end{equation}

\subsection{Segmentation Loss}
The segmentation loss $\mathcal{L}_{\text{seg}}$ ensures accurate 2D instance mask prediction for the target object. It consists of a weighted sum of Binary Cross-Entropy (BCE) and Dice losses:
\begin{equation}
    \mathcal{L}_{\text{seg}} = w_{\text{bce}} \mathcal{L}_{\text{bce}} + w_{\text{dice}} \mathcal{L}_{\text{dice}}
\end{equation}
where $w_{\text{bce}}=2.0$ and $w_{\text{dice}}=0.5$. Let $\hat{M}_{i}$ be the predicted mask logit in frame $i$ and $M_{i}$ be its ground-truth. The components are defined as:
\begin{equation}
    \mathcal{L}_{\text{bce}} = -\frac{1}{N} \sum_{i=1}^{N} \frac{1}{HW} \sum_{j=1}^{H \times W} \left[ M_{i,j} \log(\sigma_{i,j}) + (1 - M_{i,j}) \log(1 - \sigma_{i,j}) \right]
\end{equation}
\begin{equation}
    \mathcal{L}_{\text{dice}} = \frac{1}{N} \sum_{i=1}^{N} \left( 1 - \frac{2 \sum_{j} \sigma_{i,j} \cdot M_{i,j} + \epsilon}{\sum_{j} \sigma_{i,j} + \sum_{j} M_{i,j} + \epsilon} \right)
\end{equation}
where $\sigma_{i,j} = \sigma(\hat{M}_{i,j})$ is the sigmoid activation and $\epsilon=1.0$ is the smoothing constant.

\subsection{Geometry Loss}
The geometry loss $\mathcal{L}_{\text{geo}}$ regularizes the 3D reconstruction and camera trajectory:
\begin{equation}
    \mathcal{L}_{\text{geo}} = \mathcal{L}_{\text{points}} + \lambda_{\text{normal}}\mathcal{L}_{\text{normal}} + \lambda_{\text{conf}}\mathcal{L}_{\text{conf}} + \lambda_{\text{cam}}\mathcal{L}_{\text{cam}}
\end{equation}

\paragraph{Point Reconstruction and Scale Alignment.}
To resolve scale ambiguity, we first solve for the optimal global scale factor $s^*$:
\begin{equation}
    s^* = \underset{s \in \mathbb{R}^+}{\arg\min} \sum_{i,j} \frac{1}{z_{i,j}} \| s \hat{\mathbf{x}}_{i,j} - \mathbf{x}_{i,j} \|_1
\end{equation}
The reconstruction loss is then formulated as:
\begin{equation}
    \mathcal{L}_{\text{points}} = \frac{1}{3NHW}\sum_{i,j} \frac{1}{z_{i,j}} \| s^* \hat{\mathbf{x}}_{i,j} - \mathbf{x}_{i,j} \|_1
\end{equation}

\paragraph{Surface Normal and Confidence Loss.}
The normal loss $\mathcal{L}_{\text{normal}}$ minimizes the angular distance between predicted and GT normals, while the confidence loss $\mathcal{L}_{\text{conf}}$ supervises the reliability map $\mathbf{C}_i$ via BCE loss against a precision-thresholded target.

\paragraph{Camera Pose Loss.}
The camera loss $\mathcal{L}_{\text{cam}}$ is defined over relative poses:
\begin{equation}
    \mathcal{L}_{\text{cam}} = \frac{1}{N(N-1)} \sum_{i \neq j} \left( \mathcal{L}_{\text{rot}}(i, j) + \lambda_{trans} \mathcal{L}_{\text{trans}}(i, j) \right)
\end{equation}
where $\mathcal{L}_{\text{rot}}$ is the geodesic distance on $SO(3)$ and $\mathcal{L}_{\text{trans}} = \mathcal{H}_\delta(s^*\hat{\mathbf{t}}_{i \leftarrow j} - \mathbf{t}_{i \leftarrow j})$ is the scale-rectified Huber loss.



\end{document}